\documentclass[runningheads]{llncs}

\usepackage{eccv}

\usepackage{eccvabbrv}

\usepackage{graphicx}
\usepackage{booktabs}
\usepackage{algorithm}
\usepackage{algpseudocode}
\usepackage[accsupp]{axessibility}  % Improves PDF readability for those with disabilities.
\usepackage{wrapfig}  % add this in your preamble
\usepackage{graphicx} % if not already included
\usepackage{hyperref}
\usepackage{xcolor,colortbl}
\usepackage{booktabs}
\usepackage{pifont} % for \ding
\usepackage{changepage}
\usepackage{amsmath}   % for \numberwithin (equations)
\usepackage{chngcntr}  % for \counterwithin (figures/tables)
\newcommand{\cmark}{\ding{51}} % ✓
\newcommand{\xmark}{\ding{55}} % ✗
\usepackage{chngcntr}
\usepackage{multirow} 
\newcommand{\myparagraph}[1]{\vspace{3pt}\noindent{\bf #1}}

\usepackage{hyperref}

\usepackage{orcidlink}
\usepackage{bbding}

\begin{document}

\title{ClipTTT: CLIP-Guided Test-Time Training Helps LVLMs See Better} 

\titlerunning{ClipTTT: CLIP-Guided Test-Time Training}

\author{Mriganka Nath$^*$ \and Anurag Das$^*$ \and Jiahao Xie~\Envelope \and Bernt Schiele}

\authorrunning{ClipTTT: CLIP-Guided Test-Time Training}

\institute{Max Planck Institute for Informatics, Saarland Informatics Campus, Germany\\
\email{\{mnath, andas, schiele\}@mpi-inf.mpg.de \quad xiejiahao10@gmail.com}
}

\def\customsymbol#1{
    \ifcase\number\value{#1}
        \or*
        \or\Envelope
    \else\@ctrerr
    \fi
}

\maketitle

\renewcommand{\footnotesize}{\fontsize{8pt}{8pt}\selectfont}
\renewcommand{\thefootnote}{\customsymbol{footnote}}
\footnotetext[1]{Equal contribution. ~~\Envelope ~~Corresponding author.}

\begin{figure}[h]
\centering
\vspace{-10pt}
\includegraphics[width=\linewidth]{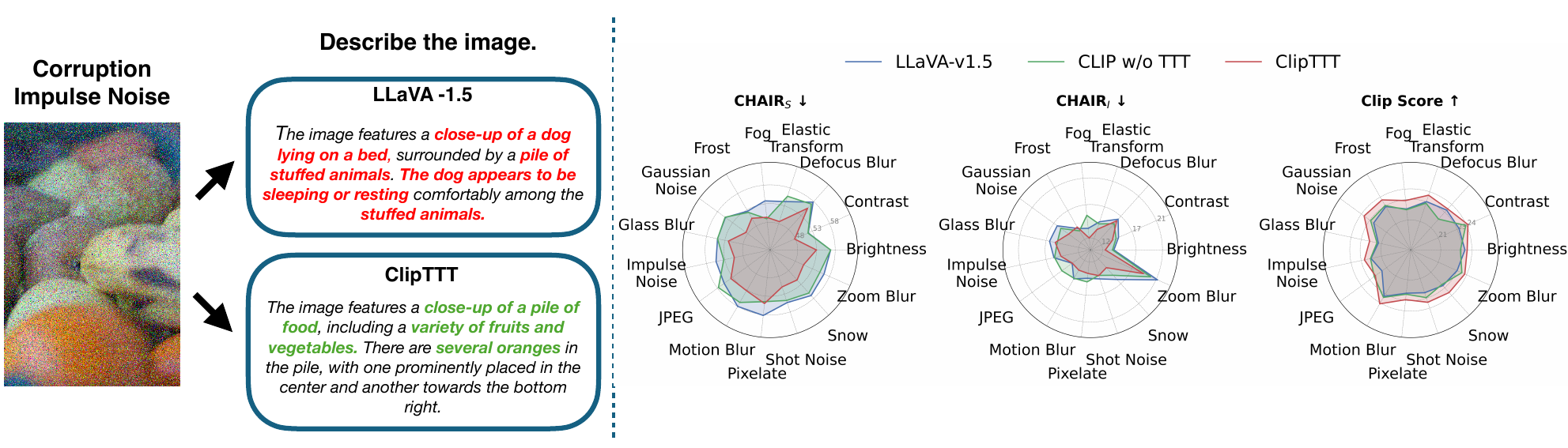}
\vspace{-20pt}
\caption{\textbf{ClipTTT improves robustness and reduces hallucinations under corruptions.} 
        \textbf{Left:} LVLMs suffer from hallucinations. Such an issue becomes worse with corruption, indicating that degradation leads to unreliable generation.
        \textbf{Right:} We benchmark the hallucinations on CHAIR~\cite{rohrbach2018object} across 15 corruptions adopted in~\cite{hendrycks2019benchmarking}. Na\"ively applying CLIP for test-time response selection may even degrade the performance. In contrast, by performing ClipTTT at test time, we consistently reduce the hallucinations of the vanilla models ($\text{CHAIR}_\text{S}$ $\downarrow$ and $\text{CHAIR}_\text{I}$ $\downarrow$).}
\label{fig:teaser}
\vspace{-25pt}
\end{figure}

\begin{abstract}
Large vision-language models (LVLMs) tend to hallucinate, especially when visual inputs are corrupted at test time. We show that such corruptions act as additional distribution shifts, significantly amplifying hallucination rates in real-world applications. To address this, we propose \textbf{CLIP}-guided \textbf{T}est-\textbf{T}ime \textbf{T}raining (\textbf{ClipTTT}), a method to adapt LVLMs under degraded conditions on the fly with a single test sample. Specifically, we leverage the image-text alignment strength of a pre-trained CLIP model as a stable guidance signal to identify reliable self-supervision targets, enabling rapid adaptation without altering the base LVLMs. Extensive experiments on standard hallucination benchmarks, with 15 common corruptions, demonstrate that ClipTTT effectively mitigates hallucinations and improves descriptive faithfulness under visual corruptions. Code will be available at \url{github.com/mrinath123/ClipTTT}.
\end{abstract}

\section{Introduction}
\label{sec:intro}

\begin{figure}[t]
\centering
\includegraphics[width=\linewidth]{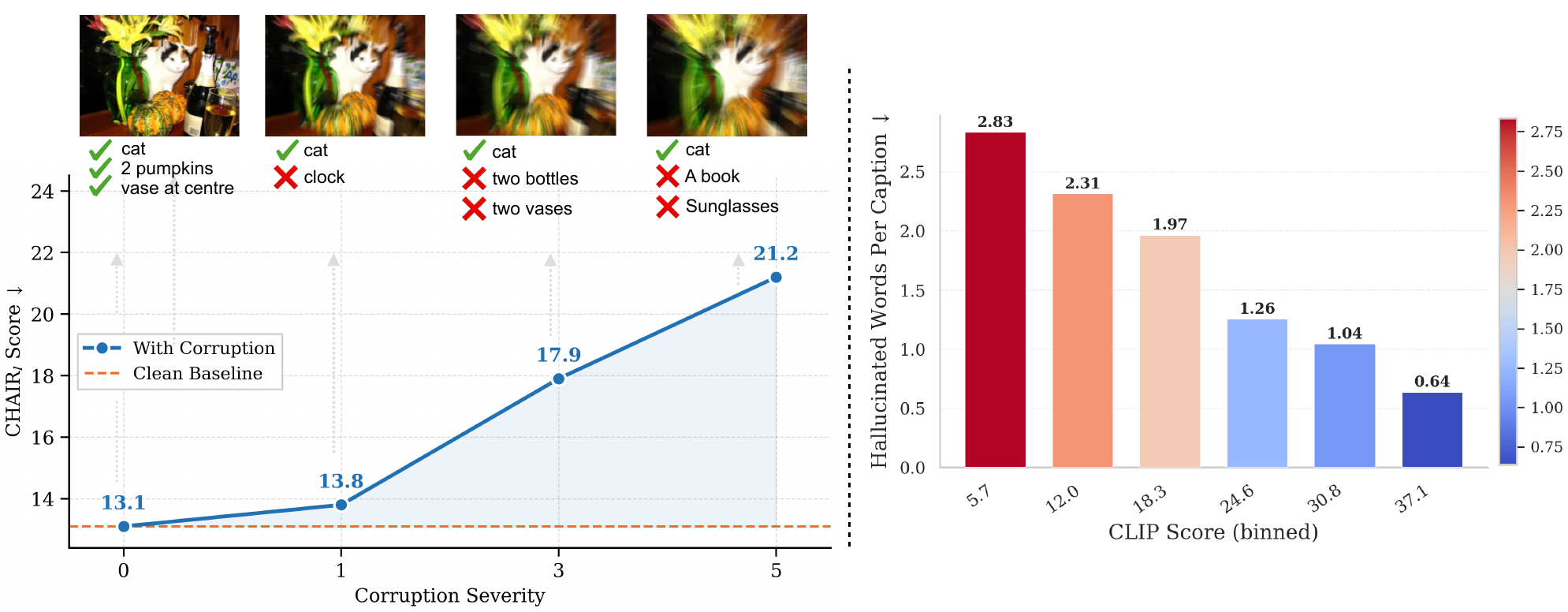}

\caption{\textbf{Left:} Hallucinations increase with corruption severities, indicating that stronger degradations reduce generation reliability (example corruption: Zoom Blur). \textbf{Right:} Captions with higher image-text alignment (CLIP Score) tend to be more factually correct. Results are averaged over 15 corruption types. Both experiments use LLaVA-v1.5-7B~\cite{liu2024improved}.}
\label{fig:teaser2}
\vspace{-12pt}
\end{figure}

Large vision-language models (LVLMs)~\cite{liu2024improved,bai2025qwen25vltechnicalreport,chen2024expanding,chenpali,zhuminigpt,alayrac2022flamingo,hurst2024gpt,bavishi2023fuyu,lillava} have achieved remarkable progress in multimodal understanding, enabling open-ended visual question answering, captioning, and reasoning.
%
% Despite these advances, hallucinations remain a fundamental reliability issue.
%
Despite these advances, hallucinations remain a major obstacle to the reliability of LVLMs' responses.
LVLMs frequently generate semantically fluent yet visually ungrounded content, undermining their trustworthiness in real-world applications. This limitation becomes particularly critical when the visual input deviates from the training distribution.
In practice, images are often affected by degradations such as noise, blur, and compression artifacts, and robustness under such conditions is essential for deployment.

Hallucinations arise from the interplay between imperfect visual grounding and strong language priors. When visual evidence is weak or ambiguous, LVLMs tend to over-rely on learned textual biases, inventing objects or wrongly describing scenes~\cite{li2023evaluating,das2026more}. Visual corruptions systematically exacerbate this failure mode by introducing additional distribution shifts at test time. We empirically establish a direct relationship between corruption severity and hallucination rate (see \cref{fig:teaser2}), demonstrating that performance progressively degrades as visual quality deteriorates. Even simple scenes trigger spurious object insertions or omissions under moderate corruptions. These findings reveal a critical robustness gap: existing LVLMs are not only sensitive to distribution shifts, but such shifts actively amplify hallucinations.

Prior efforts to mitigate hallucinations primarily rely on training-free inference-time interventions. Some approaches modify decoding strategies to encourage visual grounding~\cite{leng2024mitigating, liu2024paying}, while others incorporate external models such as CLIP to refine generated outputs~\cite{deng2024seeing}.
Although effective on clean inputs, these methods do not adapt model parameters to the corrupted test samples. As a result, they lack the capacity to compensate for image-specific degradation patterns. Robustness to unpredictable corruptions requires per-sample adaptation rather than static post-hoc correction.

In this work, we take the \textit{first step} toward mitigating hallucinations under corrupted visual inputs. We propose ClipTTT, a lightweight test-time training (TTT) framework to adapt LVLMs on the fly with a single test sample.
ClipTTT operates on a single corrupted test instance and performs rapid test-time optimization without modifying the base architecture.
Specifically, a teacher LVLM first produces multiple candidate captions for the input. We then leverage the strong image–text alignment signal of a pretrained CLIP model~\cite{radford2021learning} to select the most visually grounded candidate as a pseudo-label. A student LVLM is subsequently adapted to this pseudo-label using parameter-efficient LoRA updates~\cite{hu2022lora}. This self-guided optimization loop specializes the model to the corrupted visual context, improving grounding while preserving the original model weights outside lightweight adaptation modules.

By performing per-sample training at test time, ClipTTT enables LVLMs to recover visual faithfulness under degraded conditions. The method requires no additional supervision and avoids costly full retraining, making it practical for real-world applications.

Our main contributions are summarized as follows:

% \begin{enumerate}
\textbf{1)} We systematically show that common visual corruptions substantially amplify hallucination rates in LVLMs, exposing a critical robustness gap under distribution shifts. Furthermore, we demonstrate that existing training-free test-time interventions fail to mitigate this degradation, motivating the need for training-based solutions.

\textbf{2)} We propose ClipTTT, a test-time training framework that leverages the robust image–text alignment of a pre-trained CLIP model as a reliable guidance signal for per-sample self-correction. ClipTTT steers the LVLM toward outputs that remain faithful and image-aligned even with severe visual corruptions.

\textbf{3)} We conduct extensive experiments on standard hallucination benchmarks across 15 common corruptions. ClipTTT significantly reduces hallucinations while improves descriptive faithfulness, outperforming strong baselines with gains of \textbf{4.9 p.p. on $\text{CHAIR}_\text{S}$} and \textbf{1.6 p.p. on $\text{CHAIR}_\text{I}$} over prior state-of-the-art, while remaining robust under diverse distribution shifts.

\section{Related Work}
\myparagraph{Visual Corruption and Robustness.}

Deep models are highly sensitive to input perturbations. Early work on adversarial examples showed that small, carefully crafted distortions can fool classifiers~\cite{kurakin2018adversarial, carlini2017towards}, sparking a large body of research on adversarial robustness~\cite{bastani2016measuring, rauber2017foolbox}. Beyond adversarial settings, natural corruptions such as noise, blur, and weather distortions occur frequently and often degrade performance more severely than adversarial attacks~\cite{dodge2017study, geirhos2017comparing}. Benchmarks like ImageNet-C~\cite{hendrycks2019benchmarking} and corrupted datasets~\cite{temel2017cure, sakaridis2021acdc} highlight this fragility. More recently, domain generalization and test-time adaptation methods~\cite{wang2022continual,wang2020tent,chen2022contrastive,gao2022visual} have been proposed to improve robustness under distribution shifts. However, their impact on hallucination behaviors in LVLMs under corrupted images remains unexplored.

\myparagraph{Test-Time Training.}
An alternative way of counteracting performance degradations under distribution shifts (\eg, visual corruptions) is test-time training (TTT)~\cite{sun2020test,liu2021ttt++,bartler2022mt3,gandelsman2022test,tsai2023convolutional,xie2025test}, which unfreezes the model at test time and fine-tunes it on each incoming test sample through self-supervision.
Recently, some works~\cite{shu2022test,feng2023diverse,ma2023swapprompt,karmanov2024efficient,zanella2024test} employ TTT for CLIP models to improve their robustness.
However, none of them have studied how to mitigate hallucinations in autoregressive LVLMs with TTT.
In contrast, ours is the first work to explore TTT in the context of hallucinations in autoregressive LVLMs, where previous methods cannot be directly applied.

\myparagraph{Hallucinations in LVLMs.}
Hallucinations in LVLMs refer to generating descriptions that deviate from the visual content. This often manifests as \textit{object hallucinations}, where models describe objects not present in the image~\cite{rohrbach2018object}. Prior works attribute this issue to parametric biases from pretraining~\cite{zhang2025poison} and over-reliance on language priors~\cite{liu2024paying}. Evaluation frameworks such as CHAIR~\cite{rohrbach2018object}, POPE~\cite{li2023evaluating}, and MME~\cite{fu2025mme} quantify hallucination severity. Mitigation strategies include fine-tuning with curated datasets~\cite{qi2020reverie, yu2024hallucidoctor} and decoding adjustments~\cite{leng2024mitigating}. 

\myparagraph{Test-Time Strategies for Hallucination Mitigation.}
Test-time methods intervene at inference without retraining the base model. Contrastive decoding approaches~\cite{li2023evaluating, wang2024mitigating} suppress hallucinations by comparing token distributions under perturbed prompts or images. Attention-based strategies~\cite{liu2024paying, jiang2025devils} strengthen the role of visual tokens, while optimization-based methods such as Poison-as-Cure~\cite{zhang2025poison} learn image-specific perturbations. CLIP-guided decoding~\cite{deng2024seeing} rescales outputs based on alignment of CLIP textual and visual features. These approaches reduce hallucinations on clean inputs but do not explicitly address robustness under corrupted images.

Our work differs by focusing on hallucinations under \textit{visually corrupted} inputs, treating them as a problem of test-time robustness. To the best of our knowledge, we propose the first application of TTT for hallucination mitigation, combining visual alignment with textual guidance during inference. 

\section{Methodology}
\label{sec:method}
\begin{figure}[t]
    \centering
    \includegraphics[width=0.9\linewidth]{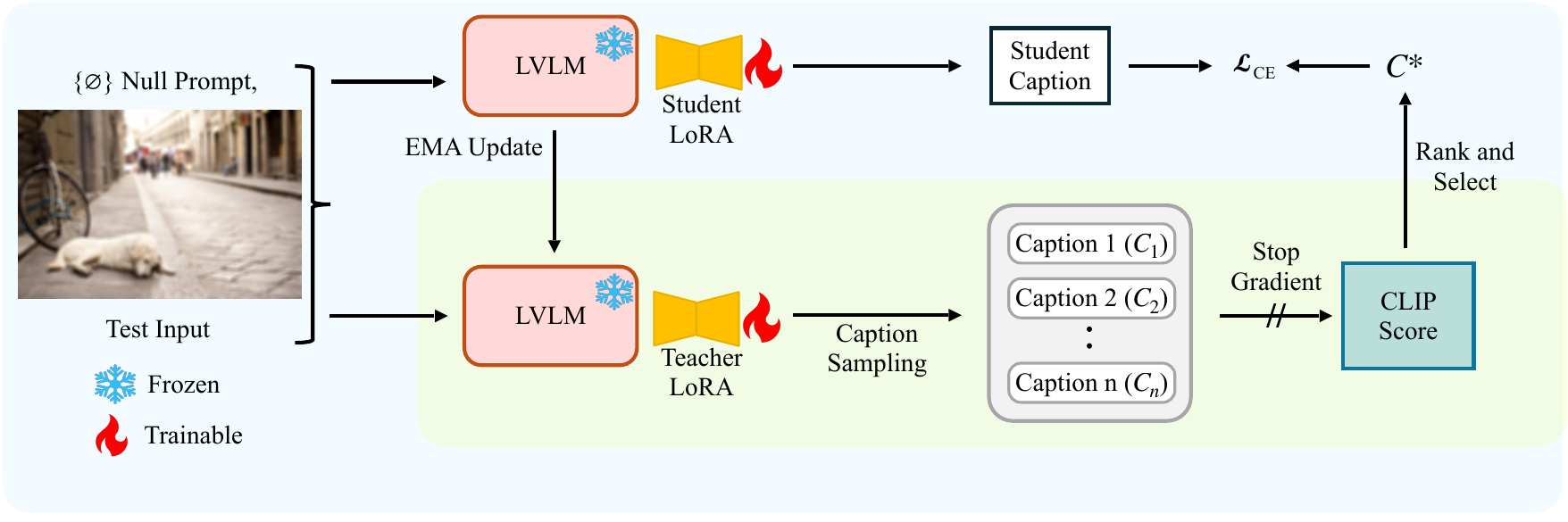}
\caption{\textbf{Overview of our ClipTTT framework.} For each single corrupted test input, we employ a student-teacher framework for on-the-fly adaptation. \textbf{(1)} The Teacher model generates $n$ diverse caption candidates via sampling. \textbf{(2)} An external CLIP model scores each candidate, and the one with the highest visual-semantic alignment is selected as the pseudo-label. \textbf{(3)} The Student model is trained for one step on this pseudo-label, with gradients updating only its parameter-efficient LoRA weights. \textbf{(4)} The Teacher's LoRA weights are then updated via an exponential moving average (EMA) of the Student's, ensuring a stable yet progressively improving training target.}
    \label{fig:method_overview}
    \vspace{-12pt}
\end{figure}

We introduce CLIP-Guided Test-Time Training (ClipTTT), a framework designed to mitigate object hallucinations in LVLMs under visually corrupted inputs. Our approach operates strictly at test time and adapts the LVLM to a single corrupted image without ground-truth supervision. It functions as a self-correction mechanism that leverages a frozen CLIP model as a supervisory signal to guide the LVLM toward visually grounded descriptions. An overview is shown in Fig.~\ref{fig:method_overview}.

\subsection{CLIP Score as a Grounding Proxy}
\label{ssec:proxy_motivation}

\myparagraph{CLIP Score.} To enable supervision without reference captions at test time, we use the CLIP Score as a proxy for factual grounding. Our key observation is that hallucinations correspond to weak image–text alignment, which CLIP is explicitly trained to measure. Given an image $x_{\text{img}}$ and caption $C$, we compute the score by averaging CLIP cosine similarities between the image and each sentence in the caption. A sentence tokenizer segments $C$ into ${s_1, \dots, s_k}$. We compute: 
\begin{equation}
    v = E_I(x_{img}) / \|E_I(x_{img})\|_2, \quad t_i \gets E_T(s_i) / \|E_T(s_i)\|_2
\end{equation}
and define CLIP Score as
\begin{equation}
   \text{CLIP Score}(x_{img}, C) =  \frac{1}{k} \sum_{i=1}^{k} (v \cdot t_i)
\end{equation}
The full procedure is summarized in~\cref{alg:clip_score_calc}.

\myparagraph{Why Sentence-Level Averaging?}
We compute similarity at the sentence level rather than over the full caption for two reasons. First, CLIP’s text encoder operates with a fixed token limit (77 tokens), and long captions may be truncated, leading to unstable similarity estimates. Sentence-level encoding avoids this issue by ensuring each segment remains within the token budget. 
Second, the whole-caption similarity can be dominated by generic high-alignment phrases, masking hallucinated statements. Sentence-level averaging penalizes visually unsupported phrases more explicitly, preventing a single aligned phrase from compensating for incorrect ones.

\myparagraph{Empirical Validation.}
We observe a clear separation between CLIP Score distributions for captions generated from clean versus corrupted images (see \cref{fig:improvement_distribution}). Furthermore, the CLIP Score correlates with hallucination rate (measured as hallucinated words per caption; see \cref{fig:teaser2}). These findings validate our central hypothesis: maximizing CLIP Score is an effective surrogate objective for reducing hallucinations.

\begin{algorithm}[h!]
\caption{CLIP Score}
\label{alg:clip_score_calc}
\begin{algorithmic}[1]
\Require Image $x_{img}$, Caption $C$, CLIP Encoders $E_I, E_T$
\State $v \gets E_I(x_{img}) / \|E_I(x_{img})\|_2$
\State $\{s_1, \dots, s_k\} \gets \text{SentenceTokenize}(C)$
\State $t_i \gets E_T(s_i) / \|E_T(s_i)\|_2,\ \forall i \in \{1,\dots,k\}$
\State $score_i \gets  (v \cdot t_i)$
\State \Return $\frac{1}{k} \sum_{i=1}^{k} score_i$
\end{algorithmic}
\end{algorithm}
\vspace{-12pt}

\subsection{CLIP-Guided Test-Time Training}
\label{ssec:tta_method}

ClipTTT is built on two principles: (1) hallucinations increase under corruptions due to weakened visual grounding, (2) CLIP provides a corruption-robust image–text alignment signal, and lightweight per-sample adaptation is sufficient to re-anchor the language model to degraded visual input. ClipTTT rationalizes these principles through stochastic exploration, CLIP-based ranking, and parameter-efficient self-training. We adapt an LVLM to a single test image $x_{\text{img}}$ via an iterative student–teacher loop. To enable efficient per-sample adaptation while keeping the base LVLM frozen, we introduce trainable Low-Rank Adaptation (LoRA) weights~\cite{hu2022lora},  $\theta_S$ and $\theta_T$. The adaptation process consists of the following steps:

\noindent\textbf{Stochastic Candidate Generation.}
At iteration $i$, the Teacher model $\mathcal{M}_T$ generates $n$ diverse captions $\mathcal{C} = {C_1, \dots, C_n}$ for the input image $x_{img}$ using stochastic decoding:
\begin{equation}
    C_j \sim \mathcal{M}_T(x_{img}, \varnothing_{\text{prompt}}), \quad \forall j \in \{1, \dots, n\}.
\end{equation}
We use a \textit{null prompt} ($\varnothing$), providing only the image as input to the LVLM. This avoids instruction-induced bias and exposes the model’s intrinsic grounding behavior. Adaptation therefore corrects hallucinations rather than prompt-specific artifacts. Further, stochastic decoding is critical, as deterministic decoding may reinforce hallucinated modes, whereas sampling enables exploration of better-grounded alternatives.

\noindent\textbf{CLIP-Guided Pseudo-Label Selection.}
Each candidate caption $C_j$ is ranked using our Sentence-Averaged CLIP Score (Algorithm~\ref{alg:clip_score_calc}). The caption with the maximum score, $C^*$, is selected as the pseudo-label for the current iteration:
\begin{equation}
    C^* = \underset{C_j \in \mathcal{C}}{\arg\max} \left( \text{CLIP Score}(x_{img}, C_j) \right).
\end{equation}
Intuitively, stochastic sampling explores the caption space, while CLIP-based ranking approximates a corruption-aware grounding objective. The selected pseudo-label acts as a high-confidence anchor with minimal hallucination.

% \paragraph{3. Student Adaptation.}
\noindent\textbf{Student Adaptation.}
The Student model's LoRA weights $\theta_S$ are updated by minimizing the cross-entropy loss $\mathcal{L}_{CE}$ between its own predictions and the pseudo-label $C^*$:
\begin{equation}
\label{eq:update}
    \theta_S^{(i+1)} \gets \theta_S^{(i)} - \eta \nabla_{\theta_S} \mathcal{L}_{CE}(\mathcal{M}_S(x_{img}, \varnothing_{\text{prompt}}), C^*),
\end{equation}
where $\eta$ is the learning rate. LoRA enables fast adaptation while preventing catastrophic drift in the pretrained LVLM, since the base weights remain frozen.

% \paragraph{4. EMA Teacher Update.} 
\noindent\textbf{EMA Teacher Update.}
The Teacher's LoRA weights $\theta_T$ are updated as an exponential moving average (EMA) of the Student's weights, governed by a decay factor $\alpha$:
\begin{equation}
    \theta_T^{(i+1)} \gets \alpha \theta_T^{(i)} + (1-\alpha) \theta_S^{(i+1)}.
\end{equation}
The EMA teacher stabilizes training by smoothing pseudo-label shifts across iterations, reducing confirmation bias from noisy samples.

\myparagraph{Adaptive Checkpoint Control.} 
Our final iteration selection and adaptive control are governed by the CLIP Score, directly optimizing for factual alignment rather than training loss. We track the student model’s CLIP Score at each iteration and select the checkpoint with the highest score for the final output. As validated in our ablation study (Sec.~\ref{sec:ablation} Tab.~\ref{tab:ablation_checkpoint}), this max-CLIP-Score strategy is empirically superior to heuristics such as selecting the final iteration or the best-loss checkpoint.

\myparagraph{Computational Overhead.} 
ClipTTT introduces minimal overhead. Adaptation involves only lightweight LoRA updates for a small number of iterations while keeping the base LVLM frozen. No architectural modifications or full retraining are required. In practice, per-sample adaptation adds modest computation overhead relative to training-free test-time approaches~\cite{deng2024seeing,zhang2025poison}, making ClipTTT practical under distribution shifts.

\section{Experiments}
\label{sec:experiments}
In this section, we empirically evaluate the effectiveness of our proposed method, ClipTTT, in mitigating hallucinations for LVLMs presented with corrupted images. We conduct a comprehensive comparison against strong baselines on standard benchmarks. Through our experiments, we aim to answer the following research questions: (1) Can our test-time training framework effectively reduce hallucinations under severe visual corruptions? (2) How does our method perform across different corruption types, baseline performance levels, and in comparison to existing hallucination reduction methods?

\subsection{Experimental Setup}
\label{ssec:exp_setup}

\begin{table*}[t]
\centering
\caption{
    \textbf{Comparison of sentence-level hallucination ($\text{CHAIR}_\text{S}$ $\downarrow$) }. Lower scores indicate better performance. 
    The results are evaluated across 15 corruption types from the CHAIR benchmark. 
    The best result in each column is in \textbf{bold}. Column headers are abbreviated.
}
\label{tab:chairs}
\resizebox{\textwidth}{!}{%
\begin{tabular}{lccccccccccccccc|c}
\toprule
\textbf{Method} & \textbf{brigh} & \textbf{cont} & \textbf{defoc} & \textbf{elast} & \textbf{fog} & \textbf{frost} & \textbf{gauss} & \textbf{glass} & \textbf{impul} & \textbf{jpeg} & \textbf{motn} & \textbf{pixel} & \textbf{shot} & \textbf{snow} & \textbf{zoom} & \textbf{AVG} \\
\midrule
Greedy Decoding & 56.8 & 51.8 & 57.8 & 54.2 & 53.8 & 52.6 & 55.4 & 55.0 & 55.4 & 55.8 & 58.0 & 58.2 & 55.4 & 57.0 & 56.4 & 55.6 \\\midrule
VCD~\cite{leng2024mitigating} & 55.6 & 50.0 & 56.6 & 51.4 & 54.2 & 53.6 & 54.0 & 54.8 & 54.6 & 57.6 & 58.8 & 55.8 & 54.0 & 55.2 & 53.2 & 54.6 \\
PAI~\cite{liu2024paying} & 58.6 & 48.6 & 56.6 & 54.6 & 51.6 & 54.0 & 56.4 & 55.6 & 56.8 & 57.2 & 62.2 & 58.2 & 54.0 & 55.2 & 57.0 & 55.8 \\
VAP~\cite{zhang2025poison} & 54.0 & 46.8 & 56.7 & 53.9 & 50.7 & 53.1 & 52.0 & 50.9 & 52.3 & 55.2 & 55.9 & 57.5 & 52.6 & \textbf{49.7} & 50.2 & 52.8 \\
CGD~\cite{deng2024seeing} & 52.4 & 51.4 & 59.1 & 56.0 & 53.8 & 52.4 & 54.0 & 56.0 & 55.4 & 57.2 & 56.1 & 59.8 & 52.2 & 55.6 & 51.6 & 54.9 \\
Restormer~\cite{zamir2022restormer} & 56.2 & 47.4 & 57.6 & 53.6 & 53.8 & 52.8 & 57.2 & 54.0 & 59.4 & 56.8 & 58.6 & 57.0 & 57.6 & 54.2 & 55.2 & 55.4 \\
NL-Means~\cite{buades2005non} & 57.6 & \textbf{42.0} & 53.2 & 56.8 & 54.2 & 55.0 & 55.0 & 57.2 & 54.0 & 58.8 & 60.8 & 57.8 & 55.0 & 55.8 & 52.4 & 55.0 \\
\midrule
 \cellcolor[HTML]{DFF6F8} ClipTTT (Ours) & \cellcolor[HTML]{DFF6F8} \textbf{48.0} & \cellcolor[HTML]{DFF6F8}   44.4 &  \cellcolor[HTML]{DFF6F8}   \textbf{50.8} & \cellcolor[HTML]{DFF6F8}  \textbf{45.0} & \cellcolor[HTML]{DFF6F8}  \textbf{44.8} & \cellcolor[HTML]{DFF6F8} \textbf{46.6} & \cellcolor[HTML]{DFF6F8}  \textbf{48.4} & \cellcolor[HTML]{DFF6F8}   \textbf{50.3} &  \cellcolor[HTML]{DFF6F8}  \textbf{49.6} & \cellcolor[HTML]{DFF6F8}  \textbf{46.0} & \cellcolor[HTML]{DFF6F8} \textbf{52.2} & \cellcolor[HTML]{DFF6F8}  \textbf{48.4} & \cellcolor[HTML]{DFF6F8}  \textbf{48.4} &  \cellcolor[HTML]{DFF6F8}  50.0 & \cellcolor[HTML]{DFF6F8}   \textbf{45.6} & \cellcolor[HTML]{DFF6F8}  \textbf{47.9} \\

\bottomrule
\end{tabular}%
}
\end{table*}

\begin{table*}[t]
\centering
\caption{
    \textbf{Comparison of instance-level hallucination ($\text{CHAIR}_\text{I}$ $\downarrow$)}. Lower scores indicate better performance. The best result in each column is in \textbf{bold}.}
\label{tab:chairi}
\resizebox{\textwidth}{!}{%
\begin{tabular}{lccccccccccccccc|c}
\toprule
\textbf{Method} & \textbf{brigh} & \textbf{cont} & \textbf{defoc} & \textbf{elast} & \textbf{fog} & \textbf{frost} & \textbf{gauss} & \textbf{glass} & \textbf{impul} & \textbf{jpeg} & \textbf{motn} & \textbf{pixel} & \textbf{shot} & \textbf{snow} & \textbf{zoom} & \textbf{AVG} \\
\midrule
Greedy Decoding & 13.7 & 14.8 & 16.4 & 14.6 & 13.5 & 13.9 & 16.4 & 16.5 & 15.8 & 13.6 & 15.2 & 14.5 & 14.8 & 16.2 & 21.2 & 15.4 \\\midrule
VCD~\cite{leng2024mitigating} & 13.9 & 15.2 & 17.0 & 14.4 & 13.6 & 15.5 & 16.2 & 15.3 & 15.5 & 14.5 & 15.3 & 14.2 & 14.8 & 16.2 & 20.3 & 15.5 \\
PAI~\cite{liu2024paying} & 14.3 & 13.8 & 15.5 & 14.8 & 12.9 & 15.1 & 16.1 & 16.7 & 15.7 & 14.1 & 16.1 & 14.6 & 14.5 & 15.6 & 21.0 & 15.4 \\
VAP~\cite{zhang2025poison} & 13.7 & 14.3 & 14.4 & 15.3 & 14.0 & 15.5 & 15.4 & 15.5 & 16.3 & 14.7 & 15.7 & 15.2 & 15.4 & 14.7 & 19.7 & 15.3 \\
CGD~\cite{deng2024seeing} & 14.1 & 14.6 & 15.5 & 15.2 & 13.9 & 15.1 & 15.9 & \textbf{14.3} & 15.5 & 13.7 & 15.6 & 15.4 & 16.0 & 15.6 & 19.6 & 15.3 \\
Restormer~\cite{zamir2022restormer} & 14.7 & 16.7 & 17.2 & 15.5 & 14.1 & 15.1 & 24.4 & 16.9 & 26.6 & 14.3 & 15.8 & 14.4 & 22.0 & 15.0 & 22.0 & 17.7 \\
NL-Means~\cite{buades2005non} & 15.9 & 15.0 & 17.5 & 16.1 & 18.7 & 18.1 & 16.9 & 16.8 & 15.1 & 14.6 & 17.7 & 14.9 & 15.4 & 17.1 & 22.6 & 16.8 \\
\midrule
\cellcolor[HTML]{DFF6F8}  ClipTTT (Ours) & \cellcolor[HTML]{DFF6F8}  \textbf{11.7} &  \cellcolor[HTML]{DFF6F8}  \textbf{13.1} &  \cellcolor[HTML]{DFF6F8}  \textbf{14.3} & \cellcolor[HTML]{DFF6F8}  \textbf{12.2} &  \cellcolor[HTML]{DFF6F8} \textbf{11.8} & \cellcolor[HTML]{DFF6F8}  \textbf{13.5} & \cellcolor[HTML]{DFF6F8}  \textbf{14.4} & \cellcolor[HTML]{DFF6F8}  15.1 &  \cellcolor[HTML]{DFF6F8}  \textbf{14.7} & \cellcolor[HTML]{DFF6F8}  \textbf{12.1} & \cellcolor[HTML]{DFF6F8}  \textbf{14.6} &  \cellcolor[HTML]{DFF6F8} \textbf{12.0} & \cellcolor[HTML]{DFF6F8}  \textbf{13.5} & \cellcolor[HTML]{DFF6F8}  \textbf{13.9} & \cellcolor[HTML]{DFF6F8}  \textbf{17.9} & \cellcolor[HTML]{DFF6F8}  \textbf{13.7} \\

\bottomrule
\end{tabular}%
}
\vspace{-12pt}
\end{table*}

\myparagraph{Corruptions.}
To rigorously test robustness, our primary evaluation uses the LLaVA-v1.5-7B~\cite{liu2024improved} model on images subjected to 15 common corruptions from~\cite{hendrycks2019benchmarking}. These cover noise (e.g., Gaussian, shot), blur (e.g., motion, defocus), weather (e.g., fog, frost), and digital artifacts (e.g., JPEG, brightness). For all experiments, we use the highest degradation level (severity 5) that is present in the Imagenet-C settings to simulate a challenging distribution shift. For each test instance, an image is first corrupted, and then the respective method~\cite{hendrycks2019benchmarking} is applied to generate a response. The results presented in our main tables (\cref{tab:chairi} and \cref{tab:chairs}) cover all 15 corruption types, providing a comprehensive measure of overall robustness.

\myparagraph{Implementation Details.}
For our test-time adaptation, we employ LoRA to adapt for each image. We apply LoRA with a rank of 8 and a scaling factor of 16 to the query and value matrices of the LLM's attention layers. The adaptation process for each image runs for 70 iterations. For efficiency, instead of every iteration, stochastic candidate generation is performed every 20 iterations. We use the AdamW optimizer and a cosine annealing learning rate scheduler with an initial learning rate of 5e-5. To generate pseudo-labels, the teacher model samples 16 candidate captions using temperature 0.8 and top-p 0.9 to form diverse captions. Its weights are updated from the student through EMA with a 0.999 decay. We set the token limit for the pseudo-label and the student captions to 512. For the CLIP model, we use ViT L/14. We use the trained Student LoRA for predictions. The checkpointing strategy is detailed in ~\cref{ssec:tta_method} and analyzed in~\cref{sec:ablation} (~\cref{tab:ablation_checkpoint}).

\myparagraph{Baselines.}
We compare ClipTTT against several methods. The standard Greedy Decoding on corrupted images serves as our primary baseline, representing the performance of an unmodified LVLM. We include four state-of-the-art training-free test time hallucination mitigation methods: VCD~\cite{leng2024mitigating}, PAI~\cite{liu2024paying}, VAP~\cite{zhang2025poison} and CGD~\cite{leng2024mitigating}. We also compare against two image pre-processing baselines, that first attempt to denoise the corrupted image before feeding it to the LVLM: Restormer~\cite{zamir2022restormer} and NL-Means~\cite{buades2005non}. Our experiments are conducted using the LLaVA-v1.5-7B~\cite{liu2024improved}. % CORRECTED: LLaVA-v1\_5

\myparagraph{Evaluation.}
%\myparagraph{CHAIR.} 
Our primary evaluation uses the Caption Hallucination Assessment with Image Relevance (CHAIR) benchmark~\cite{rohrbach2018object}, which detects hallucinations by comparing objects mentioned in generated captions with the ground-truth object set of the image. CHAIR reports two metrics: instance-level ($\text{CHAIR}_\text{I}$), measuring the fraction of hallucinated objects, and sentence-level ($\text{CHAIR}_\text{S}$), measuring the fraction of captions containing at least one hallucinated object. To assess generalization beyond COCO images and categories, we additionally evaluate on the NoCaps validation set~\cite{agrawal2019nocaps}, which contains images from OpenImages~\cite{kuznetsova2020open}, following ~\cite{deng2024seeing}. The dataset is divided into two subsets - Near-Domain (containing at least one COCO object) and Out-of-Domain (containing only novel objects). Finally, we perform an open-ended evaluation using an LMM judge (GPT-4o) following~\cite{liu2024paying}, assessing caption \textit{Accuracy} (image–text consistency) and \textit{Detailedness} (caption diversity). For the main results, we report performance across 15 corruption types (\cref{tab:chairs}, \cref{tab:chairi}), while ablations and analyses are conducted on Zoom Blur at maximum corruption severity.

\begin{table}[t]
\centering
\caption{\textbf{Effect of model scales and architectures.} \textbf{Left:} Scalability analysis across model sizes (1.1B to 13B). \textbf{Right:} Evaluation on diverse architectures. ClipTTT consistently reduces hallucinations across all setups.}
\small
\setlength{\tabcolsep}{3pt}
\renewcommand{\arraystretch}{0.9}

\begin{tabular}{cc}

\resizebox{0.48\linewidth}{!}{
\begin{tabular}{lcc}
\toprule
\textbf{Model} & \textbf{$\text{CHAIR}_\text{S}$ $\downarrow$} & \textbf{$\text{CHAIR}_\text{I}$ $\downarrow$}  \\
\midrule
TinyLLaVA-1.1B~\cite{zhou2402tinyllava} & 59.8 & 20.2  \\
\cellcolor[HTML]{DFF6F8}  \quad + ClipTTT & \cellcolor[HTML]{DFF6F8}  \textbf{59.4} &  \cellcolor[HTML]{DFF6F8}  \textbf{19.9} \\
\cmidrule{1-3}

LLaVA-1.5-7B~\cite{liu2024improved} & 56.4 & 21.2  \\
\cellcolor[HTML]{DFF6F8}  \quad + ClipTTT &  \cellcolor[HTML]{DFF6F8}  \textbf{45.6} & \cellcolor[HTML]{DFF6F8}  \textbf{18.0} \\
\cmidrule{1-3}

LLaVA-1.5-13B~\cite{liu2024improved} & 55.8 & 20.2  \\
\cellcolor[HTML]{DFF6F8}  \quad + ClipTTT & \cellcolor[HTML]{DFF6F8}  \textbf{52.2} & \cellcolor[HTML]{DFF6F8}  \textbf{15.1}  \\
\bottomrule
\end{tabular}
}
&
\resizebox{0.48\linewidth}{!}{
\begin{tabular}{lcc}
\toprule
\textbf{Model} & \textbf{$\text{CHAIR}_\text{S}$ $\downarrow$} & \textbf{$\text{CHAIR}_\text{I}$ $\downarrow$} \\
\midrule

InstructBLIP-7B~\cite{dai2023instructblip} & 45.8 & 17.4 \\
\cellcolor[HTML]{DFF6F8}  \quad + ClipTTT & \cellcolor[HTML]{DFF6F8}  \textbf{45.3} &  \cellcolor[HTML]{DFF6F8}  \textbf{16.9}  \\
\cmidrule{1-3}

InternVL2-2B~\cite{chen2024expanding} & 49.3 & 16.6  \\
\cellcolor[HTML]{DFF6F8}  \quad + ClipTTT & \cellcolor[HTML]{DFF6F8}   \textbf{49.1} & \cellcolor[HTML]{DFF6F8}  \textbf{16.1} \\
\cmidrule{1-3}

Qwen2-VL-2B~\cite{wang2024qwen2vlenhancingvisionlanguagemodels} & 43.5 & 19.1  \\
\cellcolor[HTML]{DFF6F8}  \quad + ClipTTT & \cellcolor[HTML]{DFF6F8}   \textbf{40.2} &  \cellcolor[HTML]{DFF6F8}  \textbf{18.7}  \\
\cmidrule{1-3}

Qwen2.5-VL-7B~\cite{bai2025qwen2} & 42.4 & 18.0 \\
\cellcolor[HTML]{DFF6F8}  \quad + ClipTTT & \cellcolor[HTML]{DFF6F8}  \textbf{35.4} &  \cellcolor[HTML]{DFF6F8} \textbf{17.9}  \\

\bottomrule
\end{tabular}
}

\end{tabular}

\label{tab:combined_architecture}
\end{table}

\begin{table}[t]
\centering
\caption{\textbf{Left: NoCaps (Near-Domain) results.} CHAIR score for NoCaps data for objects present in COCO. \textbf{Right: Nocaps (Out-of-Domain) results.} CHAIR score for NoCaps data for objects beyond in COCO. ClipTTT demonstrates strong generalization by lowering hallucination rates on unseen categories.
}

\scriptsize
\setlength{\tabcolsep}{2pt}
\renewcommand{\arraystretch}{1.05}

\begin{tabular}{cc}

\resizebox{0.48\linewidth}{!}{
\begin{tabular}{lcc}
\toprule
\textbf{Method} & \textbf{$\text{CHAIR}_\text{S}$ $\downarrow$} & \textbf{$\text{CHAIR}_\text{I}$ $\downarrow$}  \\
\midrule
Greedy Decoding & 63.3 & 24.0  \\
VAP~\cite{zhang2025poison} & 51.5 & 21.0 \\
VCD~\cite{leng2024mitigating} & 51.8 & 23.9 \\\midrule
\cellcolor[HTML]{DFF6F8}  ClipTTT (Ours) & \cellcolor[HTML]{DFF6F8}  \textbf{50.0} & \cellcolor[HTML]{DFF6F8}  \textbf{19.8} \\
\bottomrule
\end{tabular}
}
&
\resizebox{0.48\linewidth}{!}{
\begin{tabular}{lccc}
\toprule
\textbf{Method} & \textbf{$\text{CHAIR}_\text{S}$ $\downarrow$} & \textbf{$\text{CHAIR}_\text{I}$ $\downarrow$}  \\
\midrule
Greedy Decoding  & 64.2 & 34.0  \\
VAP~\cite{zhang2025poison} & 56.6 & 31.2 \\
VCD~\cite{leng2024mitigating} & 56.0 & 34.4 \\\midrule
\cellcolor[HTML]{DFF6F8}  ClipTTT (Ours) & \cellcolor[HTML]{DFF6F8}  \textbf{49.5} & \cellcolor[HTML]{DFF6F8}  \textbf{30.2}  \\
\bottomrule
\end{tabular}
}

\end{tabular}

\label{tab:nocaps_combined}
\vspace{-12pt}
\end{table}

\subsection{Results}
\label{ssec:main_results}
\myparagraph{Comparison on CHAIR.} 
Tables~\ref{tab:chairs} and \ref{tab:chairi} present the average results on the CHAIR benchmark across all 15 corruptions. Our method, ClipTTT, demonstrates a clear and consistent superiority over all baselines. On average, ClipTTT achieves the lowest (best) $\text{CHAIR}_\text{S}$ score of 47.9 and the lowest $\text{CHAIR}_\text{I}$ score of 13.7, indicating a significant reduction in both sentence-level and object-level hallucinations. Notably, while pre-processing denoisers like Restormer~\cite{zamir2022restormer} and NL-Means~\cite{buades2005non} can sometimes improve results on specific corruptions, they often catastrophically fail on others (e.g., NL-Means~\cite{buades2005non} on `contrast' for $\text{CHAIR}_\text{S}$), leading to poor average performance. In contrast, our per-sample adaptation approach is robust across all corruption types.

\myparagraph{Generalization Across Model Scales and Architectures.}
\label{sec:more_models}
While our primary analysis focuses on LLaVA-1.5-7B~\cite{liu2024improved}, we further evaluate the robustness of ClipTTT across different model scales and architectural designs to ensure our findings are not specific to a single backbone. To assess scalability, we extend our evaluation to the larger LLaVA-1.5-13B~\cite{liu2024improved} as well as the lightweight TinyLLaVA-1.1B~\cite{zhou2402tinyllava}. Furthermore, to demonstrate applicability across diverse LVLM architectures, we benchmark ClipTTT on InstructBLIP-7B~\cite{dai2023instructblip}, InternVL2-2B~\cite{chen2024expanding}, Qwen2-VL-2B-Instruct~\cite{wang2024qwen2vlenhancingvisionlanguagemodels} and Qwen2.5-VL-7B-Instruct~\cite{bai2025qwen2}. As shown in Tab.~\ref{tab:combined_architecture}, ClipTTT consistently mitigates hallucinations across all tested models, demonstrating that our test-time training framework is both model-agnostic and scalable.

\begin{table}[t]
\centering
\scriptsize
\setlength{\tabcolsep}{2pt}
\renewcommand{\arraystretch}{1.05}

\caption{\textbf{Left: Ablation on checkpoint strategy.} CHAIR score based on which checkpoint is used for inference. \textbf{Right: Ablation on teacher update strategy.} CHAIR score based on how the teacher model is updated.}

\begin{tabular}{cc}
\resizebox{0.48\linewidth}{!}{
\begin{tabular}{lcc}
\toprule
\textbf{Method} & \textbf{$\text{CHAIR}_\text{S}$ $\downarrow$} & \textbf{$\text{CHAIR}_\text{I}$ $\downarrow$}  \\
\midrule
Final Iteration      & 46.2 & 18.4  \\
Best Loss Iteration  & 48.2 & 19.2  \\
\midrule
\cellcolor[HTML]{DFF6F8}  Ours (Max CLIP Score) & \cellcolor[HTML]{DFF6F8}  \textbf{45.6} & \cellcolor[HTML]{DFF6F8}  \textbf{17.9} \\
\bottomrule
\end{tabular}
}
&
\resizebox{0.48\linewidth}{!}{
\begin{tabular}{lcc}
\toprule
\textbf{Method} & \textbf{$\text{CHAIR}_\text{S}$ $\downarrow$} & \textbf{$\text{CHAIR}_\text{I}$ $\downarrow$}  \\
\midrule
Fixed Teacher   & 53.9 & 18.1  \\
Dynamic Teacher & 52.0 & 18.7 \\
\midrule
\cellcolor[HTML]{DFF6F8}  Ours (EMA Teacher) & \cellcolor[HTML]{DFF6F8}  \textbf{45.6} & \cellcolor[HTML]{DFF6F8}  \textbf{17.9}\\
\bottomrule
\end{tabular}
}

\end{tabular}
% \vspace{-20pt}
\label{tab:ablation_checkpoint}
\end{table}

\begin{table}[t]
\centering

\caption{\textbf{Left: Ablation on LoRA module placement.} CHAIR score for different placements of LoRA training. \textbf{Right: Ablation on pseudo-label quality.} CHAIR score compared between the baseline, the CLIP ranked pseudo-label before TTT starts, and ClipTTT. Our CLIP-based selection provides a better starting point than the baseline, which is further improved by TTT.}

\scriptsize
\setlength{\tabcolsep}{4pt}
\renewcommand{\arraystretch}{1.05}

\begin{tabular}{cc}

\resizebox{0.48\linewidth}{!}{
\begin{tabular}{lcc}
\toprule
\textbf{Method} & \textbf{$\text{CHAIR}_\text{S}$ $\downarrow$} & \textbf{$\text{CHAIR}_\text{I}$ $\downarrow$}  \\
\midrule
Vision Encoder LoRA & 48.0 & 18.1  \\
Vision + LLM LoRA   & 45.8 & 18.0  \\
\midrule
\cellcolor[HTML]{DFF6F8}  Ours (LLM LoRA only) & \cellcolor[HTML]{DFF6F8}  \textbf{45.6} & \cellcolor[HTML]{DFF6F8}  \textbf{17.9}  \\
\bottomrule
\end{tabular}
}
&

\resizebox{0.48\linewidth}{!}{
\begin{tabular}{lcc}
\toprule
\textbf{Method} & \textbf{$\text{CHAIR}_\text{S}$ $\downarrow$} & \textbf{$\text{CHAIR}_\text{I}$ $\downarrow$}  \\
\midrule
Greedy Decoding & 56.4 & 21.2  \\
CLIP w/o TTT & 55.2 & 20.1 \\
\midrule
\cellcolor[HTML]{DFF6F8}  Ours (ClipTTT) & \cellcolor[HTML]{DFF6F8}  \textbf{45.6} & \cellcolor[HTML]{DFF6F8}  \textbf{17.9}  \\
\bottomrule
\end{tabular}
}

\end{tabular}
%\vspace{-20pt}
\label{tab:ablation_lora_pseudo_combined}
\end{table}

\myparagraph{Generalization Beyond COCO.}
\label{sec:nocaps_eval}
To evaluate generalization beyond COCO~\cite{lin2014microsoft}, we test on 500 images from the Nocaps validation set~\cite{agrawal2019nocaps}, sourced from OpenImages~\cite{kuznetsova2020open}, following ~\cite{deng2024seeing}  (see~\cref{tab:nocaps_combined}). The evaluation maps 600 fine-grained classes to 90 coarse-grained categories and partitions images into \textbf{Near-Domain} (containing at least one COCO object) and \textbf{Out-of-Domain} (containing only novel objects). Across both splits, ClipTTT outperforms greedy decoding and the strongest baselines from~\cref{tab:chairi,tab:chairs}, demonstrating strong generalization beyond COCO images and object categories.

\begin{figure}[t]
    \centering
    \includegraphics[width=0.9\columnwidth]{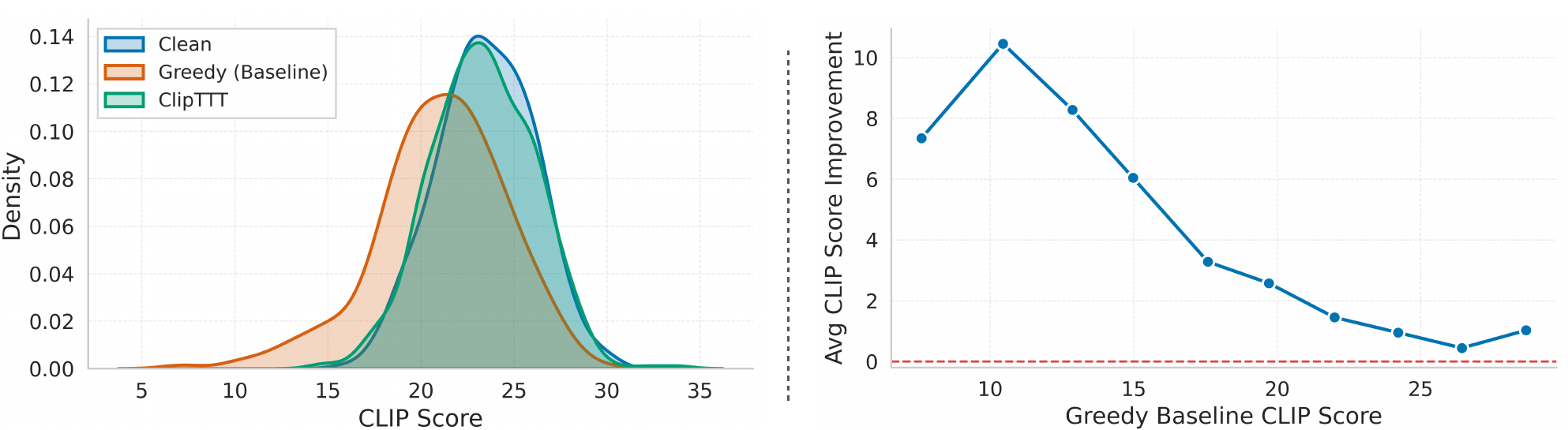} 
    \caption{\textbf{Left: CLIP Score distribution comparison.} Kernel density estimates of CLIP Scores for captions of Clean images (upper bound), corrupted images with baseline Greedy decoding, and corrupted images after applying Our Method (ClipTTT). Our method shifts the degraded distribution significantly towards the clean distribution. \textbf{Right: Improvement trend by baseline score.} The average improvement provided by ClipTTT is higher when the initial baseline caption has a low-to-medium quality score. The method adaptively reduces its intervention as the baseline quality improves.}
    \label{fig:improvement_distribution}
    \vspace{-15pt}
\end{figure}
\myparagraph{LLM Based Evaluation.}
To complement CHAIR, we conduct an open-ended evaluation using a state-of-the-art LMM judge. Following Liu et al.~\cite{liu2024paying}, we randomly sample 50 image–caption pairs from COCO val2014 using the same prompt as in the CHAIR benchmark. Evaluation is performed on clean images to verify that ClipTTT preserves generation quality. We use GPT-4o as the judge and evaluate two dimensions: Accuracy (image–text consistency) and Detailedness (richness of detail). We compare against LLaVA-1.5-7B~\cite{liu2024improved}, VCD~\cite{leng2024mitigating}, and VAP~\cite{zhang2025poison}, following~\cite{liu2024paying} (details in the supplement). As shown in Tab.~\ref{tab:gpt_analysis}, ClipTTT consistently produces more accurate responses than the baselines, improving Accuracy (5.3 vs 5.7) while maintaining comparable Detailedness.

\begin{figure}[t]
\centering
\includegraphics[width=\textwidth]{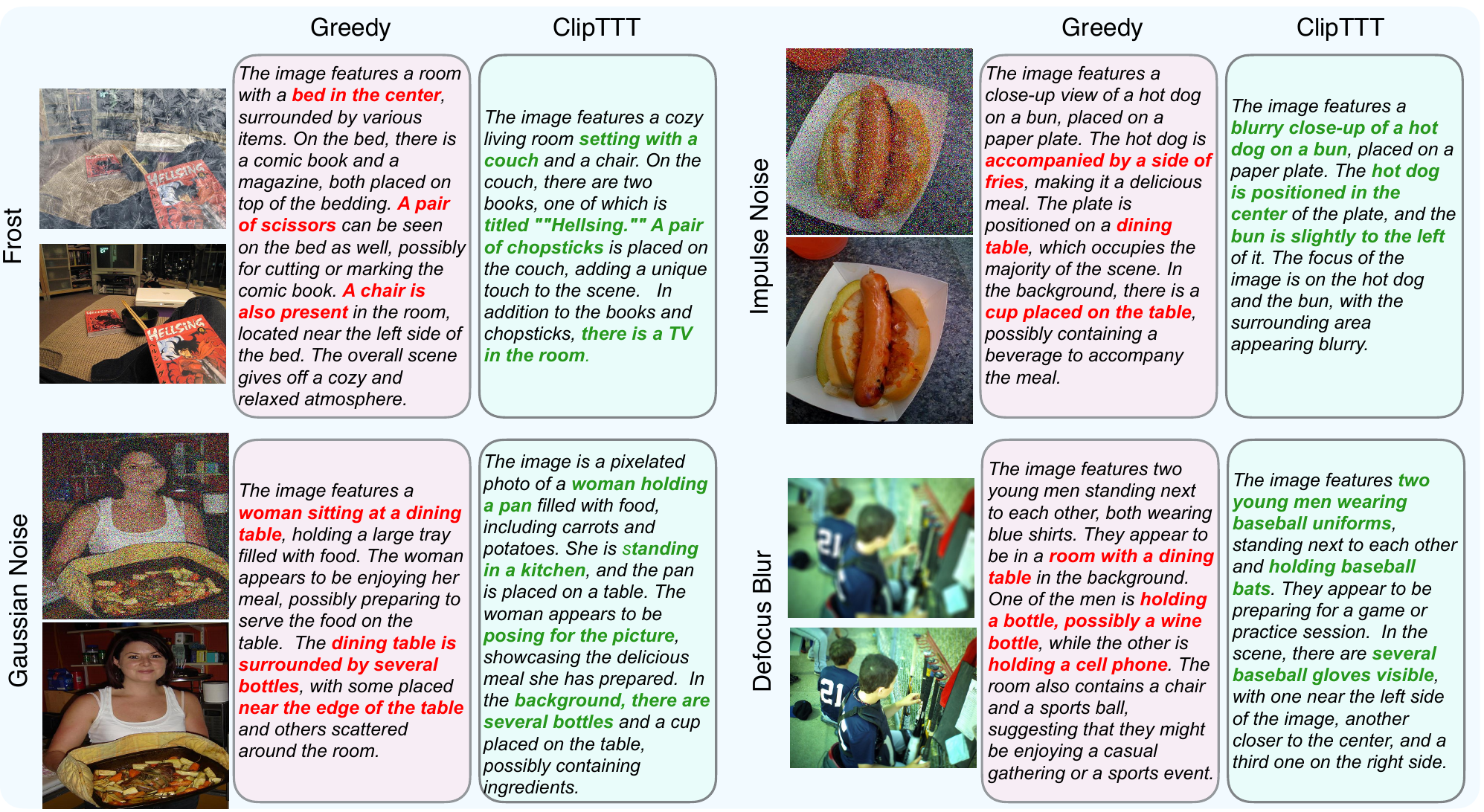}
\caption{\textbf{Qualitative Comparison.} Comparison on corrupted images with identical prompt - ``\textit{Please describe the image in detail.}''; clean images shown for reference. Corruptions (severity 5): frost, impulse noise, gaussian noise and defocus blur. \textcolor{red}{Red text} denotes hallucinations in the baseline's greedy decoding, while \textcolor{OliveGreen}{green text} highlights accurate corrections by ClipTTT. Evaluation on LLaVA-v1.5-7B~\cite{liu2024improved}. }
\label{fig:qualitative_examples}
\vspace{-20pt}
\end{figure}

\myparagraph{Qualitative Analysis.}
Figure~\ref{fig:qualitative_examples} shows that ClipTTT mitigates hallucinations across diverse corruptions. The baseline exhibits object hallucination under impulse noise and frost (inventing a ``side of fries'' or ``scissors''), and misinterprets the scene under Gaussian noise and defocus blur (e.g., claiming a standing woman is ``sitting'' or mistaking baseball players for men holding a ``wine bottle''). In contrast, ClipTTT grounds predictions in visible content (correctly recognizing ``baseball uniforms'' and text like ``Hellsing''), correctly identifying objects and preserving scene context across corruptions, thereby reducing both object- and scene-level hallucinations.

\begin{figure}[htbp]
    \centering
    % Left minipage: Table
    \begin{minipage}[c]{0.3\linewidth}
        \centering
        \scriptsize 
        \setlength{\tabcolsep}{1pt}
        \renewcommand{\arraystretch}{1.0} 
        \begin{tabular}{l|cc}
        \toprule
        \textbf{Method} & \textbf{$\text{CHAIR}_\text{S}$ $\downarrow$} & \textbf{$\text{CHAIR}_\text{I}$ $\downarrow$}  \\
        \midrule
        \multicolumn{3}{c}{\textit{Severity 1}} \\
        \hline

        Greedy Decoding & 49.9 & 13.8  \\
        VCD~\cite{leng2024mitigating} & 49.8 & 13.4 \\
        VAP~\cite{zhang2025poison} & 47.3 & 13.5 \\
        \cellcolor[HTML]{DFF6F8}  ClipTTT (Ours) & \cellcolor[HTML]{DFF6F8}  \textbf{46.5} &  \cellcolor[HTML]{DFF6F8}  \textbf{12.9}  \\
        \midrule
        \multicolumn{3}{c}{\textit{Severity 3}} \\
        \hline
        Greedy Decoding & 50.0 & 17.9  \\
        VCD~\cite{leng2024mitigating} & 49.0 & 17.6  \\
        VAP~\cite{zhang2025poison} & 48.9 & 16.1  \\
        \cellcolor[HTML]{DFF6F8}  ClipTTT (Ours) &  \cellcolor[HTML]{DFF6F8}  \textbf{48.8} & \cellcolor[HTML]{DFF6F8}  \textbf{15.1}  \\
        \bottomrule
        \end{tabular}
        \quad
    \end{minipage}%
    \hfill
    \begin{minipage}[c]{0.5\linewidth}
        \centering
        \includegraphics[width=\linewidth]{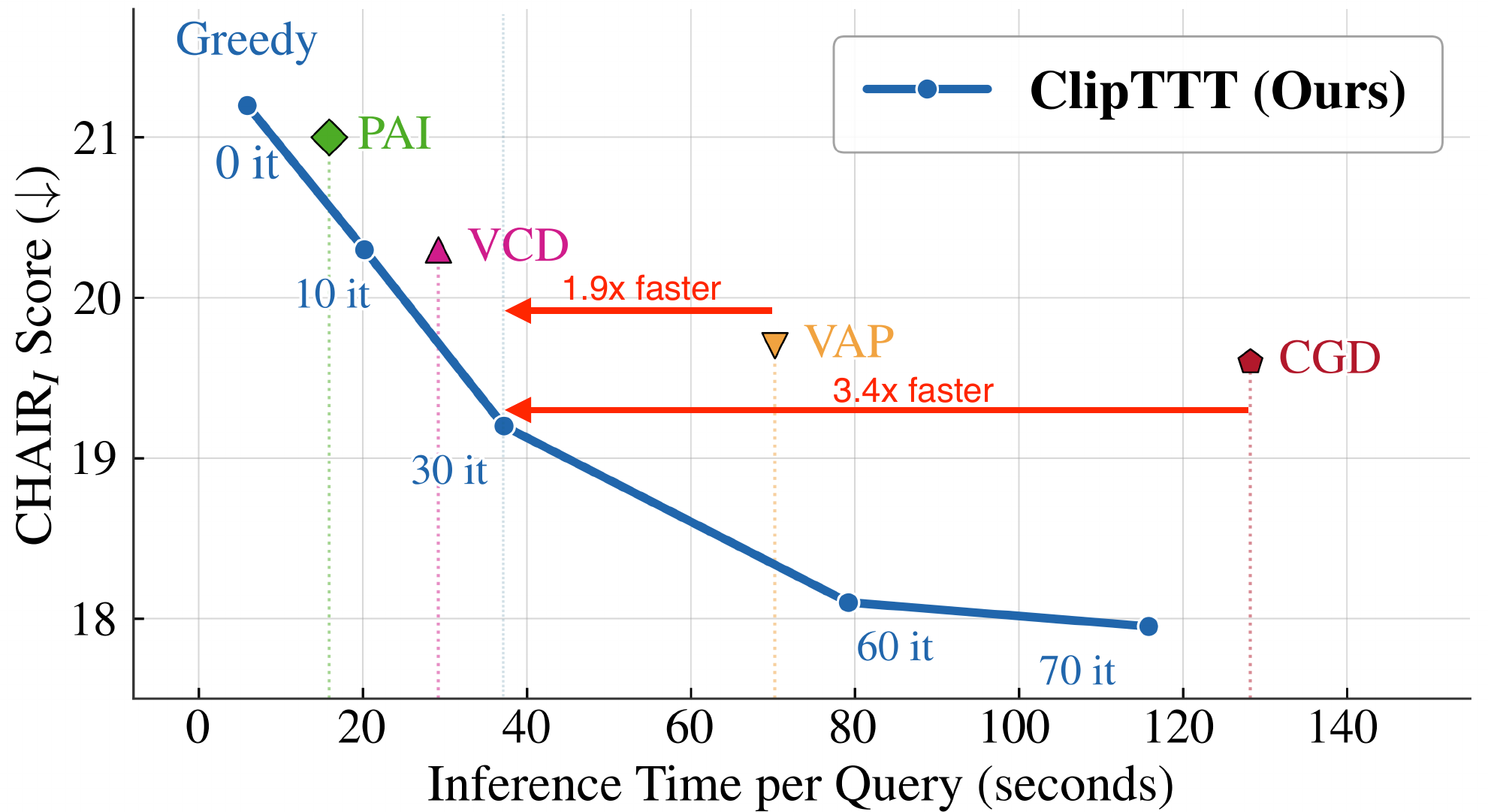}
    \end{minipage}

    \caption{ 
\textbf{Left: Ablation on corruption severity.} Sev. 1 and Sev. 3 CHAIR scores for Greedy Decoding, training-free test-time methods (VAP~\cite{zhang2025poison}, VCD~\cite{leng2024mitigating}), and ClipTTT.  
\textbf{Right: Inference time vs. performance tradeoff.} Tradeoff curve of ClipTTT at different iterations and against different test-time approaches.}
    \label{fig:ablation_severity}
    \vspace{-20pt}
\end{figure}

\subsection{Ablations and Analysis}
\label{sec:ablation}
\myparagraph{Checkpoint Strategy.}
During test-time training, selecting the right model iteration for prediction is critical. Since we lack ground-truth labels, we compare three strategies: using the \textit{final iteration}, the iteration with the \textit{best loss}, or our proposed method of selecting the iteration with the \textit{maximum CLIP Score}. Table~\ref{tab:ablation_checkpoint} shows that maximizing the self-supervised CLIP Score yields the best performance across all metrics, validating it as a powerful signal for identifying the optimal checkpoint on a per-image basis.

\myparagraph{Teacher Update Strategy.}
We investigate our EMA-based teacher update by comparing it to a ``Fixed Teacher" (the pseudo-label is fixed after the first iteration) and a ``Dynamic Teacher" (weights copied from the student per iteration). Table~\ref{tab:ablation_checkpoint} shows the EMA Teacher is superior for reducing hallucinations. A fixed teacher is sub-optimal, while a dynamic teacher is too unstable. The EMA strategy strikes a crucial balance, allowing the teacher to improve smoothly and provide progressively better supervision.

\myparagraph{LoRA Module Placement.}
We evaluate different placements of LoRA modules to determine where adaptation is most effective. As shown in Tab.~\ref{tab:ablation_lora_pseudo_combined}, applying LoRA only to the LLM achieves the best results. Using LoRA on the vision encoder or on both the vision encoder and LLM, leads to worse performance. This aligns with our setting, where the vision features remain mostly stable under corruption, and the main benefit comes from adjusting the LLM to better interpret these features.

\myparagraph{Quality of Initial Pseudo-Labels.}
We analyze the quality of the pseudo-labels, CLIP w/o TTT, selected via CLIP Score before adaptation begins. Table~\ref{tab:ablation_lora_pseudo_combined} shows this initial selection alone provides a significant improvement over the greedy baseline ($\text{CHAIR}_\text{S}$ 56.4 vs. 55.2), confirming CLIP's utility as a strong proxy for semantic fidelity. Our full TTT process then further refines this strong starting point, achieving a final $\text{CHAIR}_\text{S}$ score of 45.6. This highlights the distinct, complementary contributions of both our selection and adaptation stages.
\begin{table}[htbp]
\centering

\caption{\textbf{Left: Average CLIP Score improvement.} CLIP Score comparison between Greedy, pseudo-label, responses before TTT, and ClipTTT. \textbf{Right: GPT-4o assisted evaluation.} Accuracy reflects the absence of hallucinations, while Detailedness
reflects the richness of the description. ClipTTT improves accuracy and maintains detailedness across methods.}

\scriptsize
\setlength{\tabcolsep}{2pt}
\renewcommand{\arraystretch}{1.1}

\begin{tabular}{cc}

\resizebox{0.48\linewidth}{!}{
\begin{tabular}{lc}
\toprule
\textbf{Method} & \textbf{Avg. CLIP Score $\uparrow$} \\
\midrule
Clean Images (Upper Bound) & 23.7 \\
\midrule
Greedy Decoding & 22.5 \\
CLIP w/o TTT    & 22.9 \\
\midrule
\cellcolor[HTML]{DFF6F8}  ClipTTT (Ours)  &  \cellcolor[HTML]{DFF6F8}  \textbf{23.6} \\
\bottomrule
\end{tabular}
}
&
\resizebox{0.48\linewidth}{!}{
\begin{tabular}{lcc}
\toprule
\textbf{Method} & \textbf{Accuracy $\uparrow$} & \textbf{Detailedness $\uparrow$} \\
\midrule
Greedy (Base) & 5.2 & 5.2 \\
\midrule
VCD~\cite{leng2024mitigating} & 4.9 & 5.1 \\
VAP~\cite{zhang2025poison}            & 5.3 & 5.2 \\
\midrule
\cellcolor[HTML]{DFF6F8}  ClipTTT (Ours) & \cellcolor[HTML]{DFF6F8}  \textbf{5.7} & \cellcolor[HTML]{DFF6F8}  \textbf{5.3} \\
\bottomrule
\end{tabular}
}

\end{tabular}
\vspace{-15pt}
\label{tab:gpt_analysis}
\end{table}

\myparagraph{CLIP Score Improvement.}
We also validate CLIP Score as an effective target by analyzing its quantitative improvement. Table~\ref{tab:gpt_analysis} shows that corruptions degrade the average CLIP Score from a clean-image ideal of 23.7 to 22.5. Our initial pseudo-label, that is CLIP w/o TTT, selection recovers a significant portion of this gap, raising the score to 22.9. The subsequent adaptation then successfully optimizes this metric further, reaching a final score of 23.6. This confirms that optimizing for CLIP Score effectively closes the visual-semantic gap caused by corruption, validating its central role in our framework.

\myparagraph{Performance Across Corruption Severities.} In~\cref{fig:ablation_severity}, we compare ClipTTT with greedy decoding and training-free test-time baselines at lower corruption severities (1 and 3). ClipTTT consistently achieves lower hallucination rates than the baselines. In particular, for Severity 1, ClipTTT reduces the average $\text{CHAIR}_\text{S}$ from 49.9 to 46.5, and for Severity 3, from 50.0 to 48.8. These results indicate that ClipTTT does not over-correct and remains effective across different corruption levels.

\myparagraph{Efficiency Analysis.}
For a fair comparison, we benchmark all methods on the same GPU (Nvidia H100), measuring the average time from image loading to final captions. As shown in the trade-off analysis (see~\cref{fig:ablation_severity}), ClipTTT allows a tunable compute budget. While our 70-step setting ($\sim$116 seconds) achieves the best robustness ($\text{CHAIR}_\text{I}$ 17.9), even 30 steps (37.2 seconds, $\text{CHAIR}_\text{I}$ 19.2) already outperforms strong training-free baselines such as VAP~\cite{zhang2025poison} ($\text{CHAIR}_\text{I}$ 19.7) and CGD~\cite{deng2024seeing} ($\text{CHAIR}_\text{I}$ 19.6). Notably, although these methods are training-free, they still incur substantial inference overhead. In contrast, ClipTTT offers a more favorable trade-off, enabling users to balance robustness and efficiency according to the available budget.

\myparagraph{Overall Distribution Shift.}
Figure~\ref{fig:improvement_distribution} visualizes the impact of our method on the distribution of CLIP Scores across clean and corrupted setting. The baseline (orange) on corrupted images shows a significant degradation, with its distribution shifted to lower scores compared to the clean images (blue). After applying ClipTTT, our method (green) successfully shifts the distribution back towards the right, closing a substantial portion of the gap to the clean image distribution. This demonstrates that our test-time adaptation effectively improves the semantic alignment between the generated text and the corrupted visual content.

\myparagraph{Adaptive Improvement Behavior.}
Figure~\ref{fig:improvement_distribution} reveals a key insight into our method's behavior: it is inherently adaptive. The y-axis shows the average improvement in CLIP Score studied before and after ClipTTT. Whereas, the x-axis shows the baseline score of the initial, un-adapted caption. ClipTTT provides the largest improvements for captions that are initially poor-to-mediocre (baseline scores between 5 and 15). For captions that are already very good (baseline score $>$ 20), our method intervenes minimally, preserving their high quality. This demonstrates that our framework intelligently applies stronger corrections when needed and avoids degrading already-faithful descriptions, making it a robust and targeted adaptation technique.

\section{Conclusion}
\label{sec:conclusion}
In this work, we have studied the hallucination issue in LVLMs, which is severely amplified by visual corruptions at test time. We introduced ClipTTT, a novel test-time training framework that performs per-instance adaptation on the fly. By leveraging the CLIP model as a robust guidance, ClipTTT identifies the most visually consistent description among multiple candidates and uses it as a pseudo-label to rapidly fine-tune lightweight adapters, effectively teaching the model to ``see better''. Our extensive evaluations in 15 challenging corruptions demonstrate that ClipTTT consistently and significantly reduces hallucination rates while improving descriptive faithfulness, outperforming previous test-time intervention methods. ClipTTT enables LVLMs to maintain reliability and factual accuracy under severe and unseen visual degradations.
We hope our explorations can pave the way for deploying LVLMs in real-world scenarios with unpredictable image quality.

\bibliographystyle{splncs04}
\bibliography{main}

\appendix
\setcounter{subsection}{0}
\renewcommand\thesubsection{\Alph{subsection}}

\renewcommand\thesection{\Alph{section}}
\numberwithin{equation}{section}
\numberwithin{figure}{section}
\numberwithin{table}{section}
\renewcommand{\thefigure}{\thesection\arabic{figure}}
\renewcommand{\thetable}{\thesection\arabic{table}}
\crefname{appendix}{Sec.}{Secs.}

\clearpage
\onecolumn

\renewcommand\thesection{\Alph{section}}

\numberwithin{equation}{section}
\counterwithin{figure}{section}
\counterwithin{table}{section}

\newcommand{\adddescription}[1]{%
\begin{adjustwidth}{0cm}{0cm}
#1
\end{adjustwidth}\vspace{1em}
}

\renewcommand{\thefigure}{\thesection\arabic{figure}}
\renewcommand{\thetable}{\thesection\arabic{table}}

\crefname{section}{App.}{Apps.} % or Sec./Secs.
\title{ClipTTT: CLIP-Guided Test-Time Training Helps LVLMs See Better} 
\institute{}
\author{}
\maketitle
{\begin{center}
\Large\bf
\phantom{skip}\\[.25em]
 \bigskip
 
{Appendix}\\[1em]
\end{center}
}
\newcommand{\additem}[2]{%
\item[\textbf{(\ref{#1})}] 
    \textbf{#2} \dotfill\makebox{\textbf{\pageref{#1}}}
}

\newcommand{\addsubitem}[2]{%
% \vspace{.2em}
    \textbf{(\ref{#1})}\hspace{1em}
    #2\\[.1em] %\dotfill\makebox{\textbf{\pageref{#1}}}
}

{\hspace{-2em}\bf\large Table of Contents\\[1em]}
In this appendix, we provide additional details and results for our CLIP-guided test-time training approach for reducing hallucinations, including:
\\[1em]

\begin{adjustwidth}{1cm}{1cm}
\begin{enumerate}
\additem{sup:sec:add_res}{Additional Results}
\adddescription{This section provides additional evaluations, including generalization to the full corruption suite across severity levels, comparisons using FAITHSCORE~\cite{jing2024faithscore} and F1, Meta-ClipTTT results, performance on clean images, and CLIP-score gains across corruptions.}
    
\additem{sup:sec:add_abl}{Additional Ablations}
\adddescription{This section reports additional ablations, including the null-prompt setting, sentence-level averaging, stochastic candidate generation frequency, and the contribution of individual ClipTTT components.}

\additem{sup:sec:add_det}{Additional Details}
\adddescription{This section provides further details on ClipTTT, including experiments on CLIP similarity as a proxy for factual grounding, CLIP-score distribution shifts, LLM-judge prompt design, and attention visualizations before and after ClipTTT.}
            
\additem{sup:sec:add_qual}{Additional Qualitative Results}
\adddescription{This section provides additional qualitative comparisons, including results on real-world street scenes.}

\end{enumerate}
\end{adjustwidth}

\setlength{\parskip}{.5em}
\clearpage

\subsection{Additional Results}
\label{sup:sec:add_res}

\myparagraph{Generalization Across Models.}
In~\cref{tab:combined_architecture} (main paper), we evaluate ClipTTT under Zoom Blur across model scales and architectural variants. Here, we extend this analysis to the full suite of 15 corruptions. We study scaling using TinyLLaVA-1.1B~\cite{zhou2402tinyllava} and LLaVA-1.5-7B~\cite{liu2024improved} (results in~\cref{tab:chairs} and~\cref{tab:chairi} of the main paper), and architectural diversity using InstructBLIP-7B~\cite{dai2023instructblip} and Qwen2-VL-2B-Instruct~\cite{wang2024qwen2vlenhancingvisionlanguagemodels}. As shown in~\cref{tab:avg_gen}, ClipTTT consistently reduces hallucinations across models, indicating that our test-time training framework is model-agnostic, scalable, and robust to distribution shifts.

\begin{table}[h]
\centering
\caption{\textbf{Generalization Across Models.} We report average hallucination across 15 corruptions showing the sentence-level ($\text{CHAIR}_\text{S}$ $\downarrow$) and instance-level ($\text{CHAIR}_\text{I}$ $\downarrow$) hallucination for three LVLMs. Lower is better.}
\label{tab:avg_gen}
\small
\setlength{\tabcolsep}{7pt}
\renewcommand{\arraystretch}{1.15}
\resizebox{0.95\linewidth}{!}{%
\begin{tabular}{l|cc|cc|cc}
\toprule
\multirow{2}{*}{\textbf{Method}} &
\multicolumn{2}{c|}{\textbf{TinyLLaVA-1.1B}}~\cite{zhou2402tinyllava} &
\multicolumn{2}{c|}{\textbf{InstructBLIP-7B}}~\cite{dai2023instructblip} &
\multicolumn{2}{c}{\textbf{Qwen2-VL-2B}}~\cite{bai2025qwen25vltechnicalreport} \\
\cmidrule(lr){2-3}\cmidrule(lr){4-5}\cmidrule(lr){6-7}
& $\textbf{CHAIR}_\textbf{S}$ $\downarrow$ & $\textbf{CHAIR}_\textbf{I}$ $\downarrow$
& $\textbf{CHAIR}_\textbf{S}$ $\downarrow$ & $\textbf{CHAIR}_\textbf{I}$ $\downarrow$
& $\textbf{CHAIR}_\textbf{S}$ $\downarrow$ & $\textbf{CHAIR}_\textbf{I}$ $\downarrow$ \\
\midrule
Greedy Decoding & 63.0 & 21.1 & 55.7 & 15.9 & 47.6 & 15.0 \\
VCD~\cite{leng2024mitigating} & 62.9 & 20.7 & 55.3 & 15.5 & 47.7 & 15.3 \\
VAP~\cite{zhang2025poison} & 62.9 & 21.6 & 55.0 & 15.7 & 47.8 & 15.2 \\
\midrule
\cellcolor[HTML]{DFF6F8} ClipTTT (Ours) &  \cellcolor[HTML]{DFF6F8} \textbf{62.4} & \cellcolor[HTML]{DFF6F8}  \textbf{18.7} & \cellcolor[HTML]{DFF6F8}  \textbf{54.6} & \cellcolor[HTML]{DFF6F8}  \textbf{15.3} & \cellcolor[HTML]{DFF6F8}  \textbf{47.2} & \cellcolor[HTML]{DFF6F8}  \textbf{14.2} \\
\bottomrule
\end{tabular}%
}
\end{table}

\myparagraph{Comparison with FAITHSCORE~\cite{jing2024faithscore}.}
We report FAITHSCORE~\cite{jing2024faithscore}, a reference-free measure of visual faithfulness, to complement standard overlap-based metrics. FAITHSCORE decomposes a response into atomic visual facts (entities, attributes, and relations) and verifies each fact against the image, providing a direct signal of hallucination and grounding quality. It reports two variants: \textbf{F-Score}, which measures fact-level faithfulness over all extracted atomic facts, and \textbf{F-Score$_S$}, which measures the fraction of sentences that contain no hallucinated facts. Following~\cite{yang2025improving}, we evaluate on 500 randomly sampled COCO images. As shown in~\cref{tab:faithscore}, ClipTTT outperforms strong training-free test-time baselines by a large margin, improving F-Score by +1.4 over VAP~\cite{zhang2025poison} and +3.1 over VCD~\cite{leng2024mitigating}.

\begin{table}[h]
\centering
\caption{\textbf{Comparison with FAITHSCORE}~\cite{jing2024faithscore}. We evaluate ClipTTT against greedy decoding and strong training-free test-time baselines. FAITHSCORE is a reference-free measure of visual faithfulness computed via GPT-4o-based verification~\cite{hurst2024gpt}. \textbf{F-Score} (fact-level) measures overall atomic-fact faithfulness of the response. \textbf{F-Score$_S$} (sentence-level) is the fraction of sentences that contain no hallucinated facts. Experiments on Zoom Blur.}
\label{tab:faithscore}
\small
\setlength{\tabcolsep}{5pt}
\renewcommand{\arraystretch}{1.1}
\resizebox{0.4\linewidth}{!}{%
\begin{tabular}{l|cc}
\toprule
\textbf{Method} & \textbf{F-Score $\uparrow$} & \textbf{F-Score$_{S}$ $\uparrow$} \\
\midrule
Greedy (Base) & 76.3 & 60.9 \\
\midrule
VCD~\cite{leng2024mitigating} & 80.0 & 63.2 \\
VAP~\cite{zhang2025poison} & 81.7 & 64.1 \\
\midrule
 \cellcolor[HTML]{DFF6F8} ClipTTT (Ours) & \cellcolor[HTML]{DFF6F8} \textbf{83.1} & \cellcolor[HTML]{DFF6F8}  \textbf{64.3} \\
\bottomrule
\end{tabular}%
}
\end{table}

\myparagraph{Comparison with F1 Metric.}
Alongside $\text{CHAIR}_\text{S}/\text{CHAIR}_\text{I}$, we report object-level F1 for captioning, which balances precision (whether mentioned objects are grounded in the image) and recall (whether image objects are covered by the caption), penalizing both hallucinated and omitted objects.
As shown in Tab.~\ref{tab:f1_comp}, ClipTTT consistently improves or preserves F1 across datasets: on COCO (left), it yields a large gain for LLaVA-1.5-13B~\cite{liu2024improved} (61.5$\rightarrow$67.4) and smaller but positive gains for InstructBLIP-7B~\cite{dai2023instructblip} and Intern2-VL-2B~\cite{chen2024expanding}, while remaining near-neutral for other LVLMs. 
On NoCaps~\cite{agrawal2019nocaps} (right), ClipTTT improves out-of-domain F1 (46.3$\rightarrow$47.4), indicating better object grounding and coverage beyond the COCO distribution.

\begin{table}[t]
\centering
\caption{\textbf{Left:} F1 comparison across LVLMs on COCO (\text{CHAIR} subset). \textbf{Right:} NoCaps (Out-of-Domain)~\cite{agrawal2019nocaps} F1 results. Higher is better.}
\small

\begin{minipage}[t]{0.66\linewidth}
\centering
\setlength{\tabcolsep}{6pt}
\renewcommand{\arraystretch}{1.15}
\resizebox{\linewidth}{!}{%
\begin{tabular}{l c c c}
\toprule
 & \textbf{LLaVA-1.5-7B}~\cite{liu2024improved} & \textbf{LLaVA-1.5-13B}~\cite{liu2024improved} & \textbf{InstructBLIP-7B}~\cite{dai2023instructblip} \\
\midrule
Base      & \textbf{59.7} & 61.5          & 59.0 \\
\cellcolor[HTML]{DFF6F8}  + ClipTTT & \cellcolor[HTML]{DFF6F8}  58.1          &  \cellcolor[HTML]{DFF6F8}  \textbf{67.4} & \cellcolor[HTML]{DFF6F8}  \textbf{59.1} \\
\midrule
 & \textbf{Qwen2.5-VL-7B}~\cite{bai2025qwen2} & \textbf{Qwen2-VL-2B}~\cite{wang2024qwen2vlenhancingvisionlanguagemodels} & \textbf{Intern2-VL-2B}~\cite{chen2024expanding} \\
\midrule
Base      & \textbf{53.0} & \textbf{44.7} & 51.2 \\
\cellcolor[HTML]{DFF6F8}  + ClipTTT & \cellcolor[HTML]{DFF6F8}  52.8          & \cellcolor[HTML]{DFF6F8}  44.2          &  \cellcolor[HTML]{DFF6F8}  \textbf{52.4} \\
\bottomrule
\end{tabular}%
}
\vspace{-2pt}
\end{minipage}
\hfill
\begin{minipage}[t]{0.33\linewidth}
\centering
\scriptsize
\setlength{\tabcolsep}{4pt}
\renewcommand{\arraystretch}{1.10}
\resizebox{\linewidth}{!}{%
\begin{tabular}{l c}
\cmidrule(lr){1-2}
 &\textbf{LLaVA-1.5-7B}~\cite{liu2024improved} \\
\midrule
Base & 46.3 \\
\cellcolor[HTML]{DFF6F8} + ClipTTT            & \cellcolor[HTML]{DFF6F8}  \textbf{47.4} \\
\bottomrule
\end{tabular}%
}
\end{minipage}

\label{tab:f1_comp}
\end{table}

\myparagraph{Meta-ClipTTT for Faster Inference.}
While ClipTTT adapts each image from a random LoRA initialization, we further introduce \textbf{Meta-ClipTTT}, a faster variant that learns a good LoRA initialization for fast test-time convergence. We meta-train the LoRA parameters on 4500 clean COCO images using the Reptile algorithm~\cite{nichol2018first}, where each image defines a task that simulates the ClipTTT inner-loop adaptation. Unlike standard ClipTTT, which targets corrupted images, Meta-ClipTTT is trained only on clean images and can, in principle, be learned on any image dataset (without labels).

Concretely, let $\phi$ denote the meta-initialization of the LoRA weights. Following~\cref{eq:update}, for each clean image $x_{\text{img}}$, we run $K=12$ inner-loop updates initialized from $\phi$ to obtain task-adapted weights $\phi^{(K)}$:
\begin{equation}
\phi^{(k+1)} \gets \phi^{(k)} - \eta\nabla_{\phi}
\mathcal{L}_{\mathrm{CE}}\Big(\mathcal{M}\big(x_{\text{img}},\varnothing_{\text{prompt}};\phi^{(k)}\big), C^*\Big),
\qquad \phi_\tau^{(0)}=\phi,
\end{equation}

where $C^*$ is the CLIP-selected pseudo-label obtained by ranking sampled candidates. Reptile then performs the meta-update by moving the initialization toward the task-adapted parameters:
\begin{equation}
\phi \gets \phi + \epsilon \left(\phi_\tau^{(K)} - \phi\right),
\end{equation}
with meta step size $\epsilon$.

\begin{figure}[t]
    \centering
    \includegraphics[width=\textwidth]{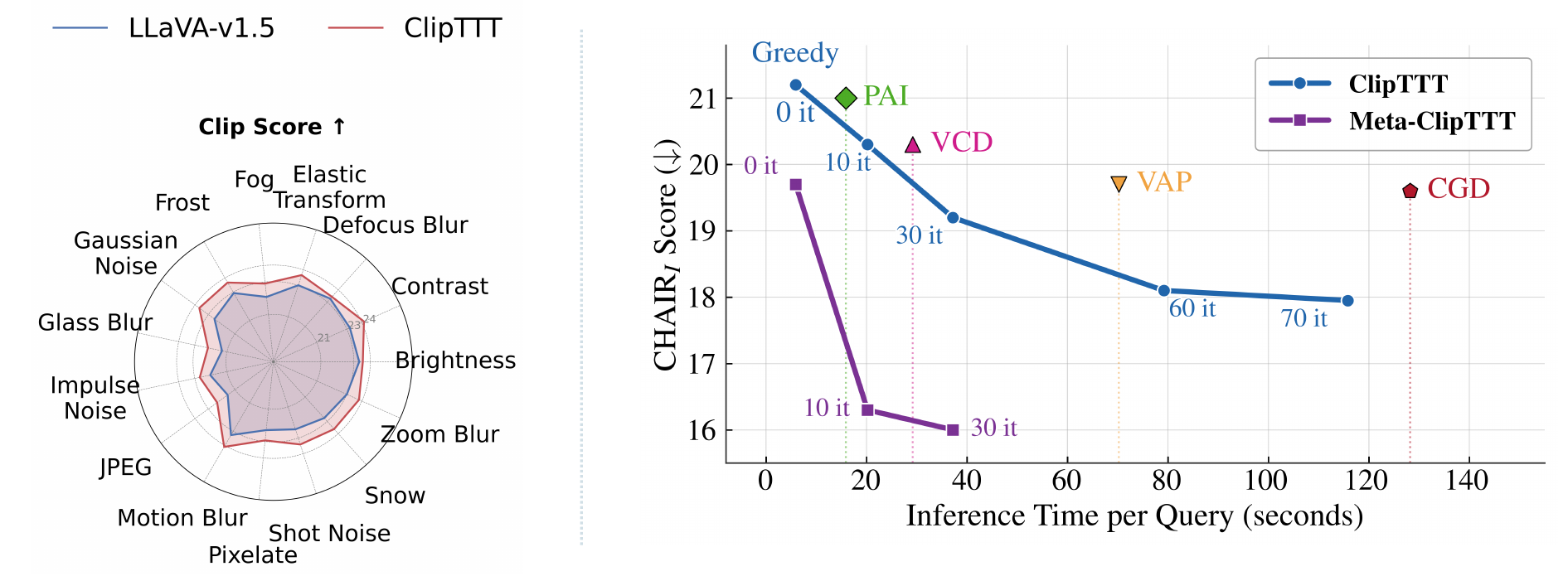} 
\caption{\textbf{Left:} CLIP Scores before and after test-time training across all corruptions. The largest gains are seen in challenging weather and noise-based corruptions like `Snow',`Gaussian Noise', and `Shot Noise'. \textbf{Right:} Meta-ClipTTT provides a stronger LoRA initialization, leading to faster convergence and improved performance. Experiment on Zoom Blur.}
\label{fig:meta_clip}
\end{figure}

Intuitively, Reptile learns an initialization that lies only a few gradient steps from a well-grounded solution under our CLIP-guided self-training objective. At test time, Meta-ClipTTT initializes both the student and teacher LoRA weights with $\phi$. As shown in Fig.~\ref{fig:meta_clip}, Meta-ClipTTT substantially reduces the number of adaptation iterations and further mitigates hallucinations, lowering inference-time cost without modifying the underlying LVLM.

\myparagraph{Comparison Across Severities.}
Our results in \cref{tab:chairs,tab:chairi} of the main paper are reported at corruption severity 5. We additionally evaluate at severities 1 and 3 to assess robustness across degradation levels. As shown in \cref{tab:avg_severity_summary}, ClipTTT consistently outperforms the greedy baseline and strong training-free test-time methods, including VAP~\cite{zhang2025poison} and VCD~\cite{leng2024mitigating}, across severities.

\begin{table}[t]
\centering
\caption{\textbf{Comparison Across Severities.} We report average hallucination across 15 corruptions showing sentence-level ($\text{CHAIR}_\text{S}$, $\downarrow$) and instance-level ($\text{CHAIR}_\text{I}$, $\downarrow$) hallucination for severity levels 1 and 3. Lower is better.}
\label{tab:avg_severity_summary}
\small
\setlength{\tabcolsep}{8pt}
\renewcommand{\arraystretch}{1.15}
\resizebox{0.75\linewidth}{!}{%
\begin{tabular}{l|cc|cc}
\toprule
\multirow{2}{*}{\textbf{Method}} &
\multicolumn{2}{c|}{\textbf{Severity 1}} &
\multicolumn{2}{c}{\textbf{Severity 3}} \\
\cmidrule(lr){2-3}\cmidrule(lr){4-5}
& $\textbf{CHAIR}_\textbf{S}$ $\downarrow$ & $\textbf{CHAIR}_\textbf{I}$ $\downarrow$
& $\textbf{CHAIR}_\textbf{S}$ $\downarrow$ & $\textbf{CHAIR}_\textbf{I}$ $\downarrow$ \\
\midrule
Greedy Decoding & 49.6 & 13.5 & 49.9 & 14.0 \\
VCD~\cite{leng2024mitigating} & 50.7 & 13.9 & 51.4 & 14.3 \\
VAP~\cite{zhang2025poison} & 49.2 & 13.4 & 50.3 & 13.8 \\
\midrule
\cellcolor[HTML]{DFF6F8} ClipTTT (Ours) &  \cellcolor[HTML]{DFF6F8} \textbf{48.4} & \cellcolor[HTML]{DFF6F8}  \textbf{12.3} & \cellcolor[HTML]{DFF6F8}  \textbf{48.8} & \cellcolor[HTML]{DFF6F8}  \textbf{12.5} \\
\bottomrule
\end{tabular}%
}
\end{table}

\myparagraph{Evaluation on Clean Images.} While ClipTTT is designed for robustness under distribution shifts, it is not restricted to corrupted images. On clean samples, ClipTTT still improves object faithfulness and reduces hallucination as shown in~\cref{tab:clean_teacher}.

\begin{table}[h]
\centering
\caption{\textbf{Left:} Clean images evaluation showing ClipTTT is not restricted to corrupted images. \textbf{Right:} Deterministic teacher ablation (stochastic teacher performs better) on Zoom Blur. Experiments on LLaVA-v1.5-7B~\cite{liu2024improved}.}
\label{tab:clean_teacher}
\scriptsize
\renewcommand{\arraystretch}{1.1}

\begin{minipage}[c]{0.4\linewidth}
\centering
\setlength{\tabcolsep}{2pt}
\resizebox{\linewidth}{!}{%
\begin{tabular}{l|cc}
\toprule
\textbf{Method} & \textbf{$\text{CHAIR}_\text{S}$ $\downarrow$} & \textbf{$\text{CHAIR}_\text{I}$ $\downarrow$} \\
\midrule
Greedy Decoding & 49.2 & 13.1 \\
VCD~\cite{leng2024mitigating}            & 49.2 & 14.1 \\
VAP~\cite{zhang2025poison}           & \textbf{47.6} & 13.9  \\
\cellcolor[HTML]{DFF6F8}  ClipTTT (Ours) & \cellcolor[HTML]{DFF6F8}  49.1 & \cellcolor[HTML]{DFF6F8}  \textbf{12.2} \\
\bottomrule
\end{tabular}%
}
\end{minipage}
\hfill
\begin{minipage}[c]{0.55\linewidth}
\centering

\setlength{\tabcolsep}{3pt}
\resizebox{\linewidth}{!}{%
\begin{tabular}{c c | c c}
\toprule
\textbf{EMA Teacher} & \textbf{Stochastic Sampling} &
\textbf{$\text{CHAIR}_\text{S}$ $\downarrow$} & \textbf{$\text{CHAIR}_\text{I}$ $\downarrow$} \\
\midrule
\xmark & \xmark & 53.9 & 18.1 \\
\cmark & \xmark & 48.2 & 20.5 \\
\cellcolor[HTML]{DFF6F8} \cmark & \cellcolor[HTML]{DFF6F8} \cmark & \cellcolor[HTML]{DFF6F8} \textbf{45.6} & \cellcolor[HTML]{DFF6F8} \textbf{17.9} \\
\bottomrule
\end{tabular}%
}

\setlength{\tabcolsep}{6pt}
\resizebox{0.9\linewidth}{!}{%
\begin{tabular}{l c c}
\toprule
 & \textbf{$\text{CHAIR}_\text{S}$ $\downarrow$} & \textbf{$\text{CHAIR}_\text{I}$ $\downarrow$} \\
\midrule
Without EMA teacher & 52.0 & 18.7 \\
\cellcolor[HTML]{DFF6F8} With EMA Teacher (Ours) & \cellcolor[HTML]{DFF6F8} \textbf{45.6} & \cellcolor[HTML]{DFF6F8} \textbf{17.9} \\
\bottomrule
\end{tabular}%
}

\end{minipage}

\end{table}

\myparagraph{Gains vs. Corruptions.} We analyze the effectiveness of ClipTTT on a per-corruption basis in Fig.~\ref{fig:meta_clip}. Our method yields a positive improvement in CLIP Score across all 15 corruption types, showcasing its broad applicability. The largest gains are seen in challenging weather and noise-based corruptions like `Snow', `Gaussian Noise', and `Shot Noise'. This highlights our method's ability to adapt and recover faithful descriptions even under severe and diverse visual distortions where baseline models struggle most.

\subsection{Additional Ablations}
\label{sup:sec:add_abl}

\myparagraph{Null Prompt.} We adopt a \textit{null prompt} ($\varnothing$), i.e., the LVLM receives only the image without any textual instruction. As discussed in Sec.~\ref{ssec:tta_method}, this reduces prompt-induced priors and isolates the model’s visual grounding behavior. We validate this choice by comparing ClipTTT under the null prompt to a fixed instruction prompt, ``\textit{Describe the image.}'' (Tab.~\ref{tab:ablation_null}). The null prompt consistently improves performance, reducing $\text{CHAIR}_\text{s}$ from 49.1 to 45.6 (lower is better), highlighting its importance for effective test-time adaptation.

\myparagraph{Sentence-Level Averaging.} As discussed in Sec.~\ref{ssec:proxy_motivation}, we compute the CLIP Score of a caption by averaging the image-text similarity across its sentences. Here, we validate this design choice by comparing sentence-level averaging against scoring the caption as a single text sequence. As shown in~\cref{tab:ablation_null}, ClipTTT with sentence-level averaging consistently outperforms the whole-caption variant, reducing $\text{CHAIR}_\text{s}$ from 49.4 to 45.6. This is particularly important because CLIP operates with a fixed context length of 77 tokens. Moreover, whole-caption scoring can be disproportionately influenced by generic, high-alignment phrases. In contrast, sentence-level averaging penalizes visually unsupported content more directly, preventing a single well-aligned phrase from masking incorrect ones.

\myparagraph{Stochastic Candidate Generation.}
For each image, we sample a set of diverse captions $\mathcal{C}=\{C_1,\dots,C_n\}$ from the teacher model $\mathcal{M}_T$. As described in Sec.~\ref{ssec:exp_setup}, we refresh these candidates every 20 iterations. We ablate this choice by varying the regeneration interval. While more frequent refresh (e.g., every iteration) is substantially more expensive (5.9 minutes per sample), it does not lead to measurable gains. As shown in~\cref{fig:gen_freq}, updating candidates every 20 iterations achieves the best performance-efficiency trade-off.

\myparagraph{Ablation on ClipTTT Components.}
In this experiment, we ablate ClipTTT to study two key design choices: the EMA teacher and stochastic sampling. We start from a variant without an EMA teacher (\textit{Fixed Teacher}), where pseudo labels are generated once at the first iteration and kept fixed throughout adaptation. We then introduce an EMA teacher to produce pseudo captions via greedy decoding, and finally replace greedy decoding with stochastic sampling to obtain diverse candidates. Adding the EMA teacher reduces hallucination ($\text{CHAIR}\text{s}$) from 53.9 to 48.2, while stochastic sampling further improves grounding, reducing $\text{CHAIR}\text{s}$ from 48.2 to 45.6. In a complementary study, we assess the importance of the EMA teacher by comparing against a variant that generates pseudo captions using the student itself. As shown in~\cref{tab:clean_teacher}, the EMA teacher consistently outperforms the student-generated alternative (45.6 vs.\ 52.0 $\text{CHAIR}_\text{s}$), highlighting the value of a more stable teacher for reliable pseudo supervision.

\begin{table}[h]
\centering
\caption{\textbf{Left:} Null-prompt ablation. ClipTTT performs test-time training without textual instructions. We compare against a prompted variant using ``\textit{Describe the image.}'' \textbf{Right:} Sentence-level similarity ablation. ClipTTT computes CLIP similarity by averaging over sentences rather than scoring the full caption. We compare against a full-caption similarity variant. Experiment on Zoom Blur.}
\label{tab:ablation_null}
\small
\setlength{\tabcolsep}{2pt}
\renewcommand{\arraystretch}{1.0}

\begin{minipage}[t]{0.48\linewidth}
\centering

\resizebox{\linewidth}{!}{%
\begin{tabular}{l c c}
\toprule
\textbf{Method} & \textbf{$\text{CHAIR}_\text{S}$ $\downarrow$} & \textbf{$\text{CHAIR}_\text{I}$ $\downarrow$} \\
\midrule
With Prompt & 49.1 & 18.5 \\
\cellcolor[HTML]{DFF6F8} With Null Prompt (Ours) & \cellcolor[HTML]{DFF6F8} \textbf{45.6} & \cellcolor[HTML]{DFF6F8} \textbf{17.9} \\
\bottomrule
\end{tabular}%
}
\end{minipage}
\hfill
\begin{minipage}[t]{0.48\linewidth}
\centering
\resizebox{\linewidth}{!}{%
\begin{tabular}{l c c}
\toprule
\textbf{Method} & \textbf{$\text{CHAIR}_\text{S}$ $\downarrow$} & \textbf{$\text{CHAIR}_\text{I}$ $\downarrow$} \\
\midrule
Full Caption & 49.4 & 19.7 \\
\cellcolor[HTML]{DFF6F8} Sentence Tokenizer (Ours) & \cellcolor[HTML]{DFF6F8} \textbf{45.6} & \cellcolor[HTML]{DFF6F8} \textbf{17.9} \\
\bottomrule
\end{tabular}%
}
\end{minipage}

\end{table}

\begin{figure}[t]
\centering

\begin{minipage}[c]{0.5\linewidth}
\centering

\small
\setlength{\tabcolsep}{6pt}
\renewcommand{\arraystretch}{1.15}
\resizebox{\linewidth}{!}{%
\begin{tabular}{l c c c}
\toprule
\textbf{Interval} & \textbf{$\text{CHAIR}_\text{S}$ $\downarrow$} & \textbf{$\text{CHAIR}_\text{I}$ $\downarrow$} & \textbf{Avg time (mins)} \\
\midrule
0 & 53.9 & 18.1 & 0.3 \\
\hline
1  & 45.1 & 17.3 & 5.9 \\
5  & 45.3 & 17.7 & 4.1 \\
20 & 45.6 & 17.9 & 1.9 \\
40 & 47.2 & 18.9 & 1.2 \\
\bottomrule
\end{tabular}%
}
\end{minipage}
\hfill
\begin{minipage}[c]{0.4\linewidth}
\centering
\resizebox{\linewidth}{!}{%
\includegraphics{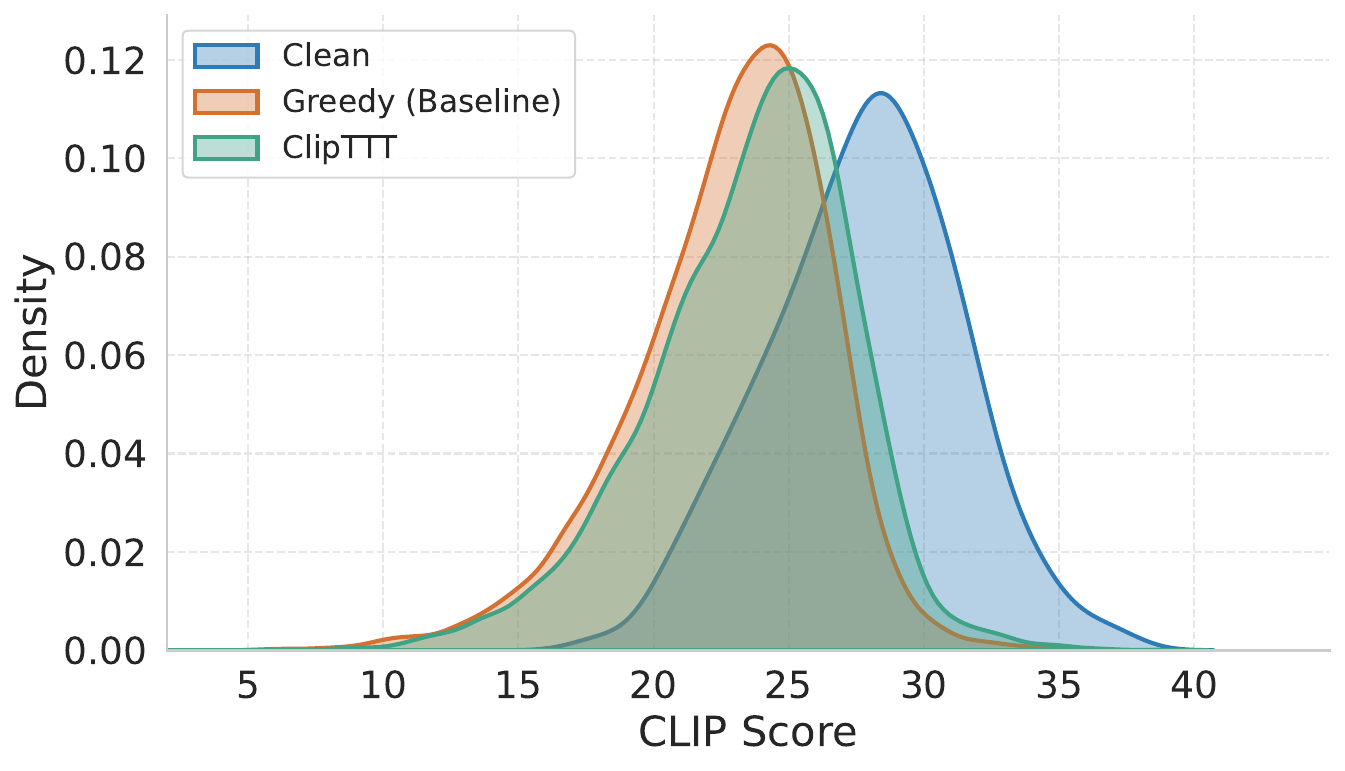}%
}
\end{minipage}
\caption{\textbf{Left:} Stochastic candidate generation frequency. Ablation of the performance-efficiency trade-off when generating candidates at different re-evaluation intervals. Interval~0 corresponds to the \textit{Fixed Teacher} setting in~\cref{tab:ablation_checkpoint}, where captions are generated once at the first iteration and then kept fixed throughout test-time training. Experiment on Zoom Blur. \textbf{Right:} CLIP Score distributions averaged over 15 corruptions.}
\label{fig:gen_freq}
\end{figure}

\begin{figure}[h]
    \centering
    \includegraphics[width=\textwidth]{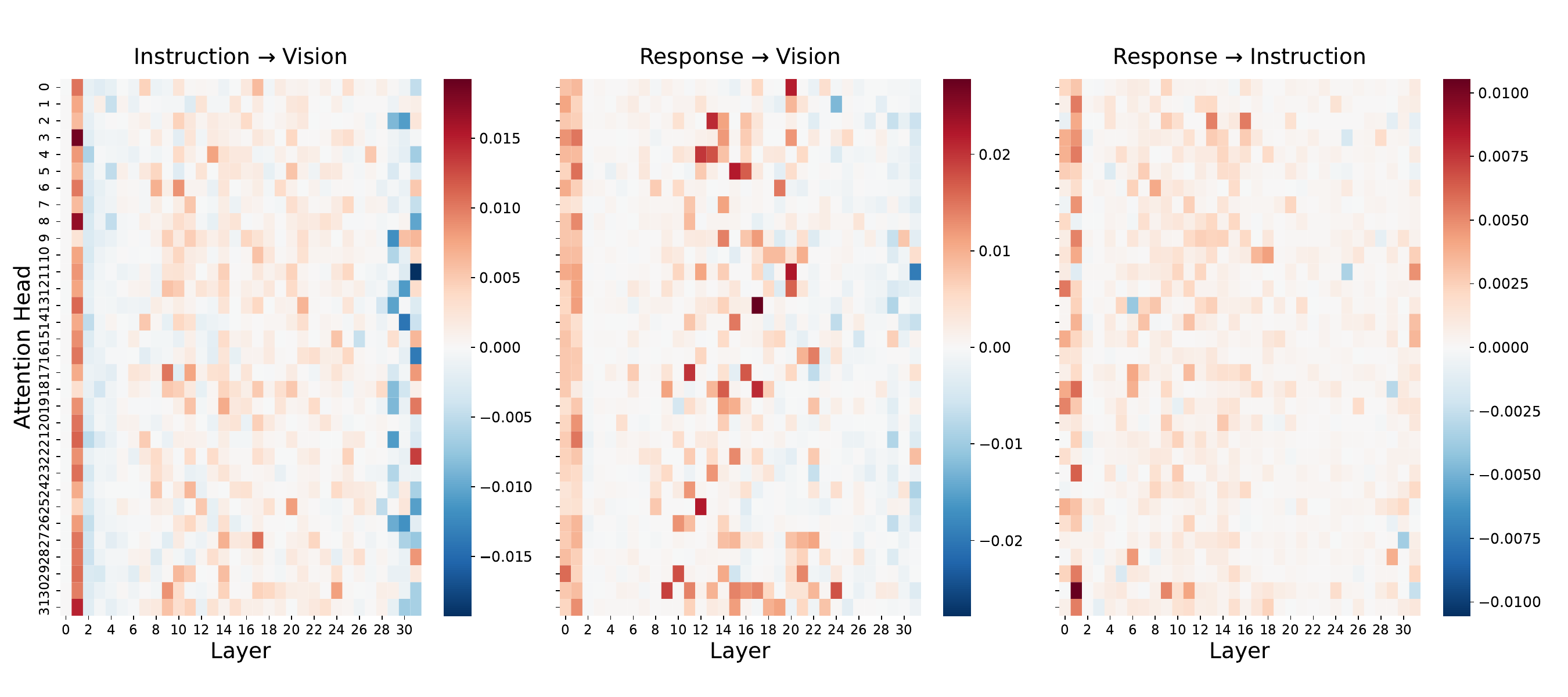} 
    \caption{\textbf{Visualization of attention weight shifts in LLaVA-1.5-7B.} We visualize the average change in attention weights ($\Delta \text{Attention} = \text{Attention}_{\text{ClipTTT}} - \text{Attention}_{\text{Base}}$). The x-axis represents the Transformer layers (0-31), and the y-axis represents attention heads (0-31). \emph{Left:} Attention from Instruction tokens to Vision tokens. \emph{Middle:} Attention from Response tokens to Vision tokens. \emph{Right:} Attention from Response tokens to Instruction tokens. Red regions indicate where ClipTTT amplifies attention, effectively countering the visual disconnect caused by corruptions. Experiments on Zoom Blur.}
\label{fig:attention_heatmap}
\end{figure}

\subsection{Additional Details.}
\label{sup:sec:add_det}

\begin{figure}[t]
    \centering
    \includegraphics[width=\textwidth]{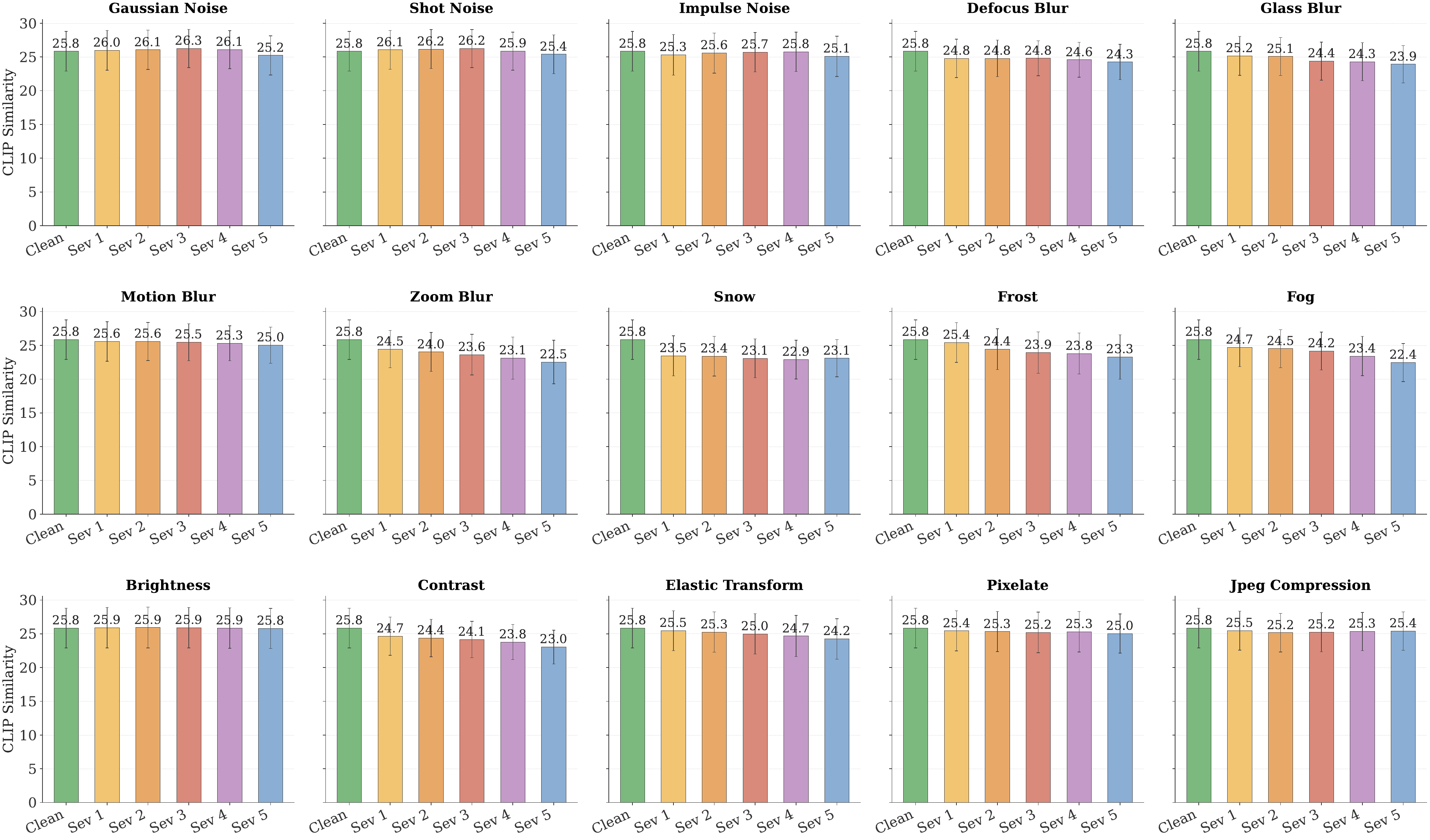} 
    \caption{\textbf{CLIP Similarity for Factual Grounding.} We compute the CLIP Score (Sec.~\ref{ssec:proxy_motivation}) between the \textit{ground-truth caption} and corrupted images across severity levels. The score remains relatively stable even under severe corruptions, motivating its use as a grounding proxy. Results are averaged over 500 COCO validation images.}
\label{fig:sim_fact}
\end{figure}

\myparagraph{CLIP Similarity for Factual Grounding.}
In~\cref{fig:teaser2} from the main paper, we observe a clear relationship between hallucination and CLIP similarity: hallucinations increase as the CLIP similarity of generated captions decreases. This motivates optimizing a CLIP-based objective to mitigate hallucinations. We further analyze the stability of CLIP similarity under corruptions by measuring the CLIP Score (Sec.~\ref{ssec:proxy_motivation}) between the \textit{ground-truth caption} and images corrupted at different severity levels. Concretely, we randomly sample 500 images from COCO-Val, apply all 15 corruptions across severity levels, and compute CLIP similarity for each corrupted image-caption pair. As shown in~\cref{fig:sim_fact}, CLIP similarity remains largely stable even at high severities for most corruptions (e.g., JPEG compression, Gaussian/Shot noise, and Brightness). These results further support the use of CLIP similarity as a robust proxy for visual grounding under distribution shift.  

\myparagraph{CLIP Score Distribution Shift.} Figure~\ref{fig:improvement_distribution} from the main paper illustrates how ClipTTT changes the distribution of CLIP Scores on clean and corrupted inputs. While this figure shows a representative example only for Zoom Blur, here we report the average distribution shift over all the 15 corruptions. As shown in~\cref{fig:gen_freq}, the baseline on corrupted images (orange) degrades substantially, with the score distribution shifting left relative to clean images (blue). After applying ClipTTT, the adapted model (green) shifts the distribution back towards right, recovering a fraction of the gap to the clean distribution. These results indicate that test-time adaptation improves semantic alignment between the generated captions and corrupted visual content.

\myparagraph{LLM Judge Prompt Design.}
To complement CHAIR, we additionally conduct an open-ended evaluation with a state-of-the-art LMM judge (Sec.~\ref{ssec:main_results}). Following~\cite{liu2024paying}, we randomly sample 50 images from COCO val2014 and generate captions using the CHAIR prompt, ``\texttt{Please describe the image in detail}.'' We then feed the judge the image and the LVLM responses, and ask it to assign 1--10 scores for Accuracy (penalizing hallucinations and any mismatch in objects, counts, positions, or colors) and Detailedness (rewarding relevant visual details while ignoring hallucinated ones), along with a brief justification to reduce ordering bias. The complete judge prompt is provided in~\cref{tab:gpt4o_prompt}. Unlike~\cite{liu2024paying}, which evaluates two LVLMs, we extend the prompt to accommodate four models. As shown in~\cref{tab:gpt_analysis} from the main paper, ClipTTT outperforms the greedy baseline and strong training-free test-time methods on both accuracy and detailedness.

\begin{table}[h]
\centering
\renewcommand{\arraystretch}{1.1} 
\caption{The prompt used for GPT-4o assisted evaluation.}
\label{tab:gpt4o_prompt}
\begin{tabular}{p{0.9\textwidth}}
\hline
\textbf{GPT-4o Prompt} \\
\hline
You are required to score the performance of four AI assistants in describing a given image. You should pay extra attention to the hallucination, which refers to the part of descriptions that are inconsistent with the image content, such as claiming the existence of something not present in the image or describing incorrectly in terms of the counts, positions, or colors of objects in the image. Please rate the responses of the assistants on a scale of 1 to 10, where a higher score indicates better performance, according to the following criteria: \\
1: Accuracy: whether the response is accurate with respect to the image content. Responses with fewer hallucinations should be given higher scores. \\
2: Detailedness: whether the response is rich in necessary details. Note that hallucinated descriptions should not count as necessary details. \\
Please output the scores for each criterion, containing only four values indicating the scores for Assistant 1, 2, 3 and 4, respectively. The four scores are separated by a space. Following the scores, please provide an explanation of your evaluation, avoiding any potential bias and ensuring that the order in which the responses were presented does not affect your judgment. \\
\\
{[Assistant 1]} \\
\{Response of Assistant 1\} \\
{[End of Assistant 1]} \\
\\
{[Assistant 2]} \\
\{Response of Assistant 2\} \\
{[End of Assistant 2]} \\
\\
\\
{[Assistant 3]} \\
\{Response of Assistant 3\} \\
{[End of Assistant 3]} \\
\\
{[Assistant 4]} \\
\{Response of Assistant 4\} \\
{[End of Assistant 4]} \\
\\
Output format: \\
Accuracy: $<$Scores of the four answers$>$ \\
Reason: \\
\\
Detailedness: $<$Scores of the four answers$>$ \\
Reason: \\
\hline
\end{tabular}
\end{table}

\myparagraph{Attention Visualization.}
To interpret the impact of ClipTTT, we analyze the changes in attention weights across Transformer layers, following~\cite{zhang2025llava}. Figure \ref{fig:attention_heatmap} visualizes the change in attention weights ($\Delta\text{Attention}$) for three interaction pathways: Instruction $\to$ Vision, Response $\to$ Vision, and Response $\to$ Instruction, averaged across 500 images, which we used for CHAIR evaluation.
The Response $\to$ Vision distribution (\emph{Middle}) is of particular interest, as prior work identifies the change of attention toward image tokens, quantified as the Visual Attention Ratio (VAR), as a primary driver of hallucinations~\cite{jiang2025devils}. Our results reveal distinct positive shifts (red regions) particularly in the middle layers in this pathway, demonstrating that ClipTTT enables the model to maintain active attention on visual embeddings throughout the decoding process, thus leading to better image-text grounding. Simultaneously, we observe increased attention in the Response $\to$ Instruction pathway (\emph{Right}), showing more alignment to textual constraints.

\subsection{Additional Qualitative Results}
\label{sup:sec:add_qual}

\begin{figure}[t]
    \centering
    \includegraphics[width=0.8\textwidth]{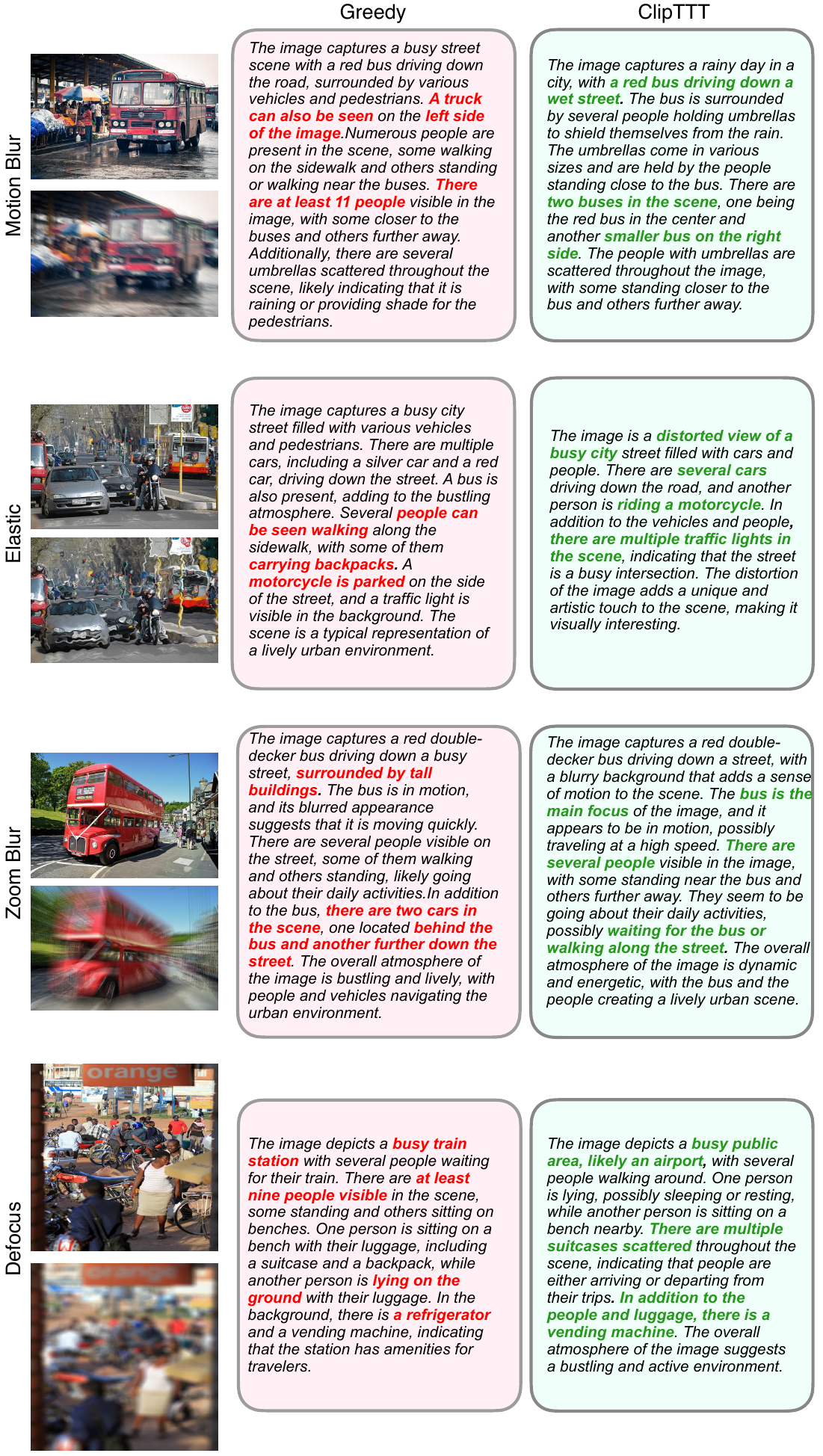} 
    \caption{Qualtiative Comparison on \textbf{Real-World Street Scenes}:  Comparison on corrupted images with identical prompt - ``\textit{Please describe the image in detail.}''; clean images shown for reference. \textcolor{red}{Red text} denotes hallucinations in the baseline's greedy decoding, while \textcolor{OliveGreen}{green text} highlights accurate corrections by ClipTTT. Evaluation on LLaVA-v1.5-7B~\cite{liu2024improved}.}
\label{fig:qual_sup_1}
\end{figure}

\begin{figure}[t]
    \centering
    \includegraphics[width=0.8\textwidth]{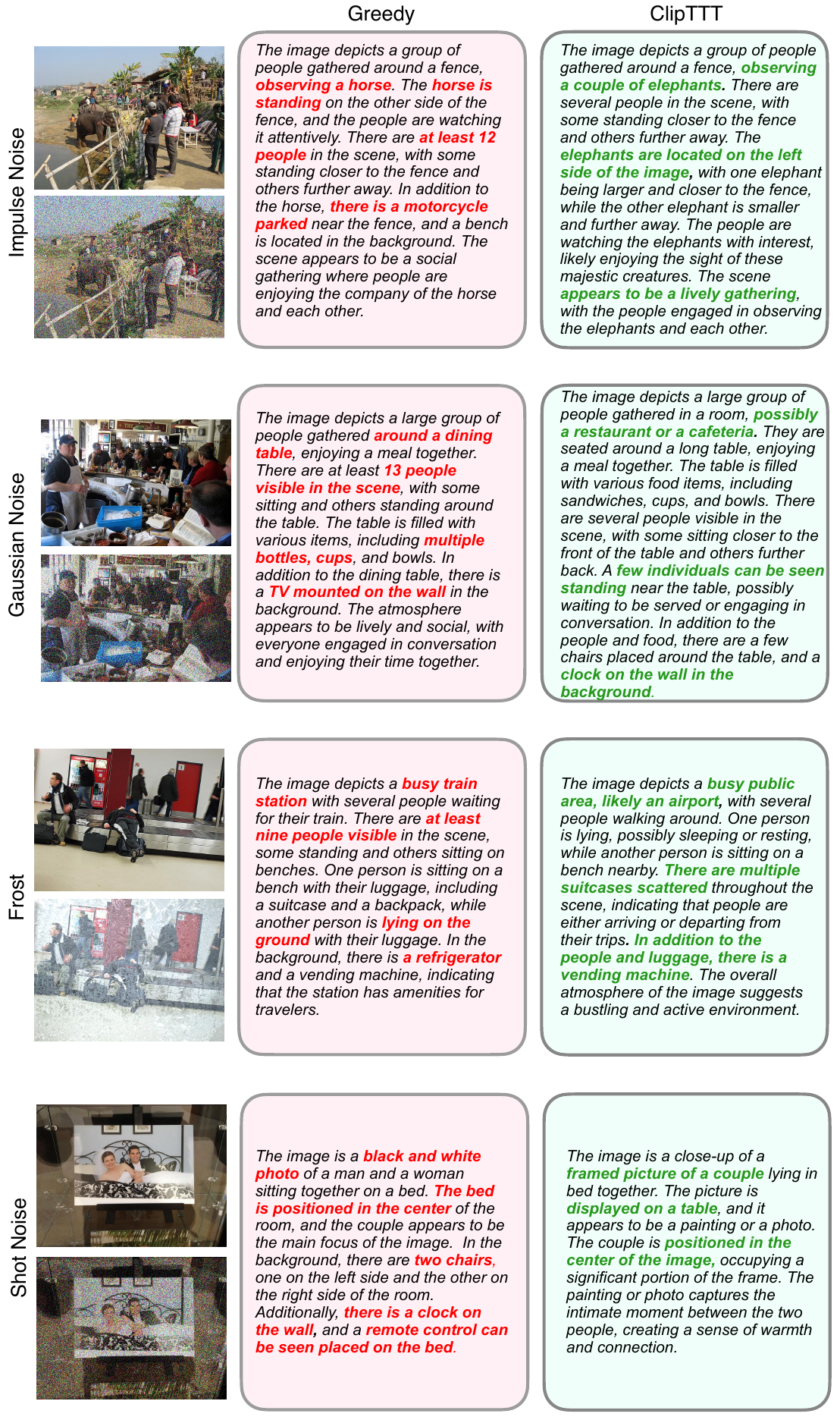} 
        \caption{Additional Qualitative Comparisons:  Comparison on corrupted images with identical prompt - ``\textit{Please describe the image in detail.}''; clean images shown for reference. \textcolor{red}{Red text} denotes hallucinations in the baseline's greedy decoding, while \textcolor{OliveGreen}{green text} highlights accurate corrections by ClipTTT. Evaluation on LLaVA-v1.5-7B~\cite{liu2024improved}.}
\label{fig:qual_sup_2}
\end{figure}

We extend the qualitative analysis with additional results on real-world street scenes in~\cref{fig:qual_sup_1}, and provide further examples in~\cref{fig:qual_sup_2}. Overall, ClipTTT consistently produces more grounded captions than the baseline. In particular, under elastic, zoom, and defocus corruptions, the baseline hallucinates content such as ``\textit{people walking}'', ``\textit{two cars in the scene}'', and ``\textit{busy train station}''. In contrast, ClipTTT improves factual correctness by generating captions that better match the visual evidence, e.g., ``\textit{person riding a motorcycle}'', ``\textit{standing near the bus}'', and ``\textit{busy public area}'' for the same inputs.

\end{document}